\documentclass[journal]{IEEEtran}
\usepackage{microtype}
\usepackage{amssymb}
\usepackage{amsmath}
\usepackage{cases}
\usepackage{algorithm}
\usepackage{algorithmic}
\usepackage{subfigure}
\usepackage{graphicx}
\usepackage{multirow}
\usepackage{mathrsfs}
\usepackage{enumitem}
\usepackage{bm}
\usepackage{amsthm}
\usepackage{mathtools}
\usepackage{booktabs} 
\usepackage{hyperref}
\usepackage{color}

\usepackage[switch]{lineno}
\usepackage{graphicx}

\ifCLASSINFOpdf
\else
\fi

\begin{document}
%
\title{Hierarchical Prototype Learning for Zero-Shot Recognition}
%
%
%

\author{Xingxing Zhang,
        Shupeng Gui,
        Zhenfeng Zhu,
        Yao Zhao,~\IEEEmembership{Senior Member,~IEEE,}
        and~Ji~Liu
\thanks{X. Zhang, Z. Zhu, and Y. Zhao are with the Institute of Information Science, Beijing Jiaotong University, Beijing 100044, China, and also with the Beijing Key Laboratory of Advanced Information Science and Network Technology, Beijing Jiaotong University, Beijing 100044, China (e-mail: zhangxing@bjtu.edu.cn; zhfzhu@bjtu.edu.cn; yzhao@bjtu.edu.cn). (\emph {Corresponding author: Yao Zhao.})}
\thanks{S. Gui and J. Liu are with the Department of Computer Science, University of Rochester, Rochester, NY 14611 USA (e-mail: shupenggui@gmail.com; jliu@cs.rochester.edu).}}


\maketitle

\begin{abstract}
Zero-Shot Learning (ZSL) has received extensive attention and successes in recent years especially in areas of fine-grained object recognition, retrieval, and image captioning. Key to ZSL is to transfer knowledge from the seen to the unseen classes via auxiliary semantic prototypes (e.g., word or attribute vectors). However, the popularly learned projection functions in previous works cannot generalize well due to non-visual components included in semantic prototypes. Besides, the incompleteness of provided prototypes and captured images has less been considered by the state-of-the-art approaches in ZSL. In this paper, we propose a \underline{h}ierarchical \underline{p}rototype \underline{l}earning formulation to provide a systematical solution (named \textbf{HPL}) for zero-shot recognition. Specifically, HPL is able to obtain discriminability on both seen and unseen class domains by learning visual prototypes respectively under the transductive setting. To narrow the gap of two domains, we further learn the interpretable super-prototypes in both visual and semantic spaces. Meanwhile, the two spaces are further bridged by maximizing their structural consistency. This not only facilitates the representativeness of visual prototypes, but also alleviates the loss of information of semantic prototypes. An extensive group of experiments are then carefully designed and presented, demonstrating that HPL obtains remarkably more favorable efficiency and effectiveness, over currently available alternatives under various settings.
\end{abstract}


\IEEEpeerreviewmaketitle

\section{Introduction}
 \IEEEPARstart{T}{raditional} object recognition tasks require the test classes to be identical or a subset of the training classes. Due to the deep learning techniques and growing availability of big data, dramatic progresses have been achieved on these tasks in recent years~\cite{he2016deep}.
 However, in many practical applications, we need the model to have the ability to determine the class labels for the object belonging to unseen classes. The following are some popular application scenarios~\cite{wang2019survey}: 
 \begin{itemize}
     \item[-] \emph{The number of target classes is large}. Generally, human beings can recognize at least 30,000 object classes. However, collecting sufficient labelled instances for such a large number of classes is challenging. Thus, existing image datasets can only cover a small subset of these classes.
     \item[-] \emph{Target classes are rare}. An example is fine-grained object recognition. Suppose we want to recognize flowers of different breeds. It is hard and even prohibitive to collect sufficient image instances for each specific flower breed. 
     \item[-] \emph{Target classes change over time}. 
     An example is recognizing images of products belonging to a certain style and brand. As products of new styles and new brands appear frequently, for some new products, it is difficult to find corresponding labelled instances. 
     \item[-] \emph{Annotating instances is expensive and time consuming}. For example, in the image captioning problem, each image in the training data should have a corresponding caption. This problem can be seen as a sequential classification problem. The number of object classes covered by the existing image-text corpora is limited, with many object classes not being covered.
 \end{itemize}
 
 \begin{figure}[t!]
\begin{center}
\centerline{\includegraphics[width=3.0in]{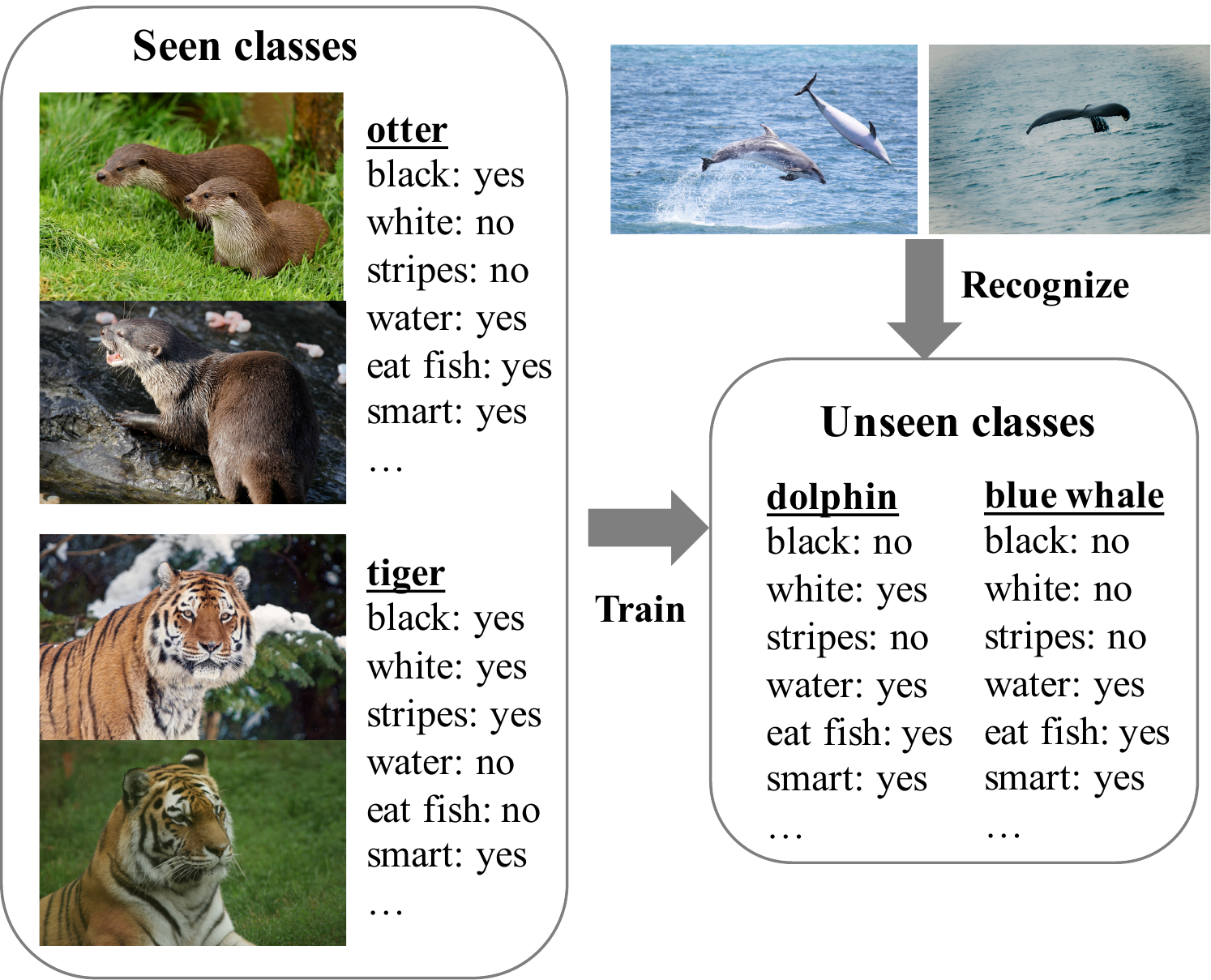}}
\caption{The visual images and semantic prototypes provided for several classes in benchmark dataset AwA2~\cite{xian2018zero}.}
\label{att_image}
\end{center}
\vskip -0.4in
\end{figure}

 To solve this problem, Zero-Shot Learning (ZSL)~\cite{mensink2014costa,zhang2015zero,wang2016zero, changpinyo2016synthesized,morgado2017semantically,xian2018feature} is proposed. The goal of zero-shot recognition is to recognize objects belonging to the classes that have no labelled samples. Since its inception, it has become a fast-developing field in machine learning with a wide range of applications in computer vision. Due to lack of labelled samples in the unseen class domain, auxiliary information is necessary for ZSL to transfer knowledge from the seen to the unseen classes. As shown in Fig.~\ref{att_image}, existing methods usually provide each class with one \emph{semantic prototype derived from text} (e.g., attribute vector~\cite{kankuekul2012online, lampert2009learning, lampert2014attribute} or word vector~\cite{akata2015evaluation, frome2013devise, socher2013zero}). This is inspired by the way human beings recognize the world. For example, with the knowledge that ``a zebra looks like a horse, and with stripes'', we can recognize a zebra even without having seen one before, as long as we know what a ``horse'' is and how ``stripes'' look like.
 
\begin{figure}[t!]
\begin{center}
\centerline{\includegraphics[width=3.0in]{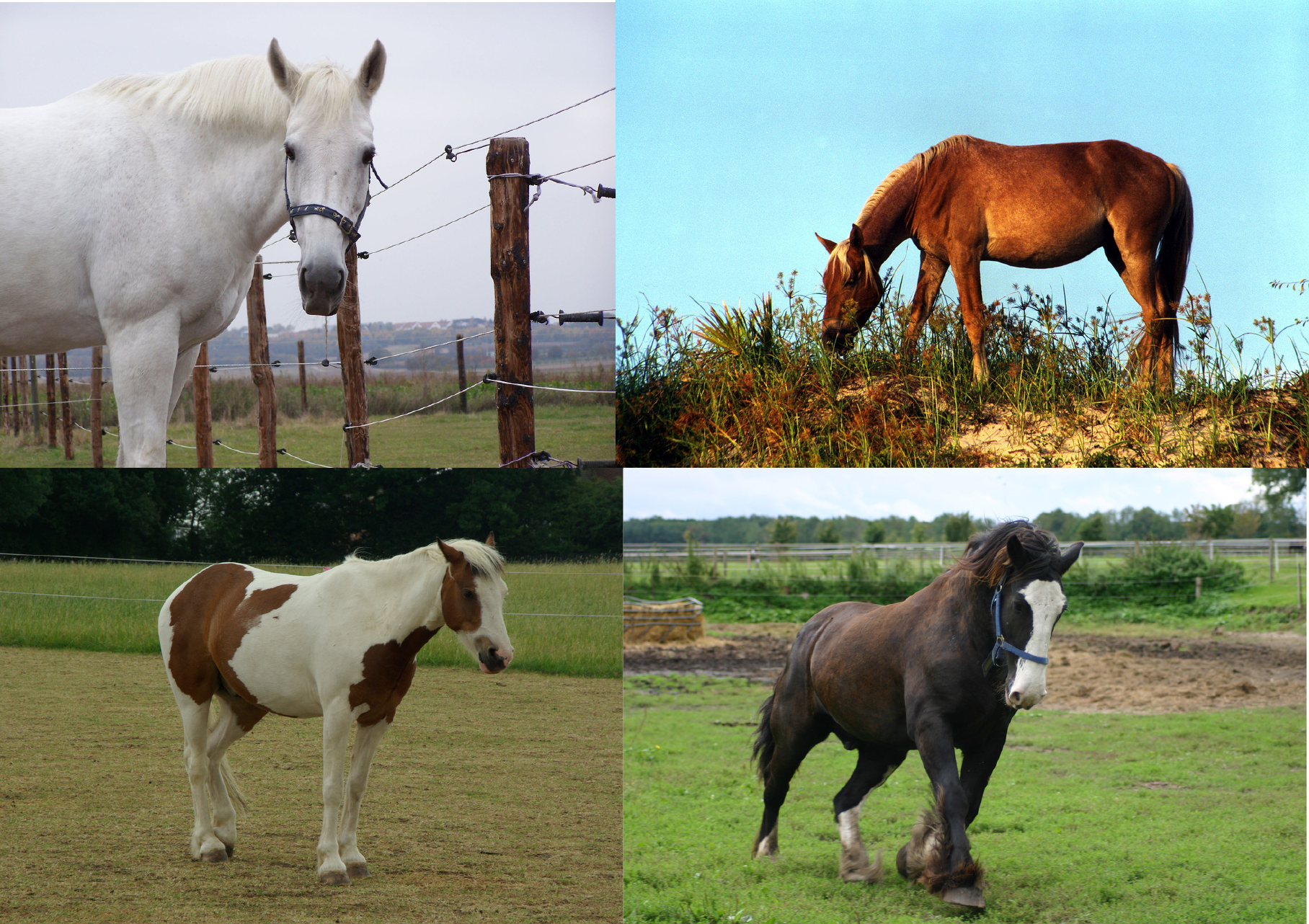}}
\caption{Several instances with different colors from the unseen class ``horse'' in benchmark dataset AwA2~\cite{xian2018zero}.}
\label{horse}
\end{center}
\vskip -0.3in
\end{figure}

\begin{figure}[t!]
\vskip -0.2in
\begin{center}
\centerline{\includegraphics[width=\columnwidth]{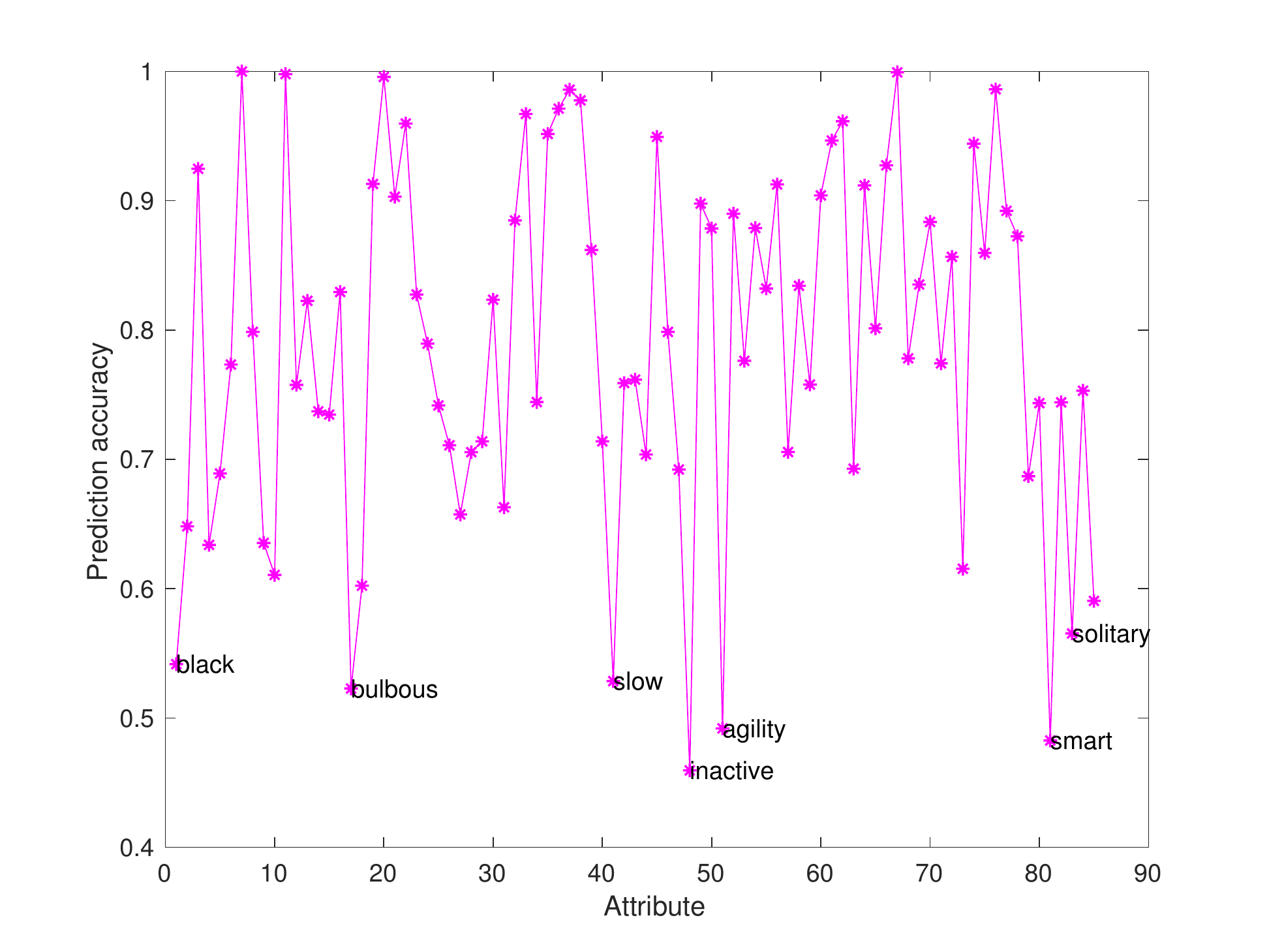}}
\caption{The predictability of each binary attribute, measured with classification accuracy by pre-trained Resnet101~\cite{he2016deep}, where we only fine-tune the last layer.}
\label{error}
\end{center}
\vskip -0.2in
\end{figure}

 Typical ZSL approaches generally adopt a two-step recognition strategy~\cite{romera2015embarrassingly,zhang2016zero,kodirov2017semantic,annadani2018preserving}. First, an image-semantics projection function is learned from the seen class domain to transfer knowledge to unseen classes. Then, the test sample is projected into the learned embedding space, where the recognition is carried out by considering the similarity between the sample and unseen classes. 
 Thus, various ZSL approaches are developed to learn a well-fitting projection function between visual features and semantic prototypes. \underline{However}, they all ignore a fact that the provided semantic prototypes are incomplete and less diversified, since both human-defined attribute vectors and automatically extracted word vectors are obtained independently from visual samples, and uniquely for each class. Consequently, the learned projection may not be effective enough to recognize the samples from the same class. For instance, there are different kinds of colors for a ``horse'' as shown in Fig.~\ref{horse}.
 \underline{Besides}, the non-visual components are often included in the provided semantic prototypes, such as ``smart'', ``agility'', and ``inactive'' in benchmark dataset AwA2~\cite{xian2018zero}. Based only on visual information, these attributes are almost impossible to be predicted and even with the level of random guess as observed from Fig.~\ref{error}. Thus, the learned projection cannot generalize well on the unseen class domain, although work normally in the seen class domain through supervised training.
 \underline{Moreover}, in practice, the visual image captured from an object cannot present all the attributes of corresponding class. 
 \textcolor{black}{As a result, a simple projection from an image to its class attribute vector is inaccurate since that image may lack some attributes (e.g., tail and claws are not captured).}
 
 \underline{Finally}, as mentioned in many ZSL approaches~\cite{rohrbach2013transfer,fu2015transductive,kodirov2015unsupervised,verma2017simple,niu2018webly,zhao2018domain,jiang2018learning}, large domain gap between the seen and unseen classes is one of the biggest challenges in ZSL. This comes from the fact that for a certain attribute (e.g., ``tail''), the unseen classes are often visually very different from the seen ones. Consequently, the projection function learned from seen-class data may not be effective enough to project an unseen object to be close to its corresponding class. To address this domain shift issue, that is, to reduce the domain distribution mismatch between the seen- and unseen- class data, a number of ZSL models resort to transductive learning~\cite{fu2015transductive,niu2018webly,zhao2018domain} by utilizing test objects from unseen classes in the training phase. 
 \textcolor{black}{Here, we also introduce a transductive setting, where the learned projection is adapted to unseen classes based on unlabelled target data. It has been proved in~\cite{fu2015transductive} that transductive ZSL models can indeed improve generalisation accuracy compared with inductive models.}
 
 For the importance of these phenomenons, we develop a novel ZSL model in this paper under the transductive setting. By learning visual prototypes and super-prototypes, instead of learning a projection between the visual and semantic spaces, the proposed model is able to avoid the aforementioned problems caused by semantic prototypes. In particular, considering the incompleteness of provided semantic prototypes and visual images, we choose to couple the semantic prototypes with the learned visual prototypes. Motivated by the fact that there exist unseen/seen prototypes that fall into the same class, we further learn the prototypes of seen and unseen prototypes, called \emph{super-prototypes}. They can bridge not only the seen and unseen class domains to induce a transductive setting, but also the visual and semantic spaces to align their structure.
 
In summary, the contributions of this work are four-fold:
\begin{itemize}
\item[-] We propose a novel transductive ZSL model (named \textbf{HPL}) which enforces discriminability on both seen and unseen class domains by learning visual prototypes respectively.
\item[-] Interpretable super-prototypes that are learned from the visual (resp. semantic) prototypes are able to bridge the two domains, since super-prototypes are shared between the seen and the unseen classes.
\item[-] By maximizing the structural consistency of visual and semantic prototypes, the representativeness of learned visual prototypes is further strengthened, thus leading to more discriminative recognition. 
\item[-] An efficient algorithm is presented to solve our model with rigorous theoretical analysis. The improvements over currently available alternatives are especially significant under various ZSL settings.
\end{itemize}

\section{Related Work}\label{Related work}
In this section, we first briefly introduce some related works on transductive ZSL, and then present a review for image-semantics projection in ZSL.

\textbf{Transductive Zero-Shot Learning.} According to whether information about the testing data is involved during model learning, existing ZSL models consist of inductive~\cite{romera2015embarrassingly,changpinyo2016synthesized,kodirov2017semantic,annadani2018preserving} and transductive settings~\cite{akata2016label,verma2017simple,zhao2018domain}. 
Specifically, this transduction in ZSL can be embodied in two progressive degrees: transductive for specific unseen classes~\cite{liu2018generalized} and transductive for specific testing samples~\cite{zhao2018domain}.
Specifically, by extending conventional ZSL into a semi-supervised learning scenario, transductive ZSL is an emerging topic recently since it can effectively rectify the domain shift caused by different distributions of the training and the test samples.
Depending on the inference strategy for the test data, existing transductive ZSL models mainly involve two groups. The first one ~\cite{rohrbach2013transfer,fu2015transductive,fu2018zero} generally constructs a graph in the semantic space, and then transfers it to the test set by label propagation. However, due to lack of labelled samples for unseen classes, such methods usually have a great probability to predict the test objects as seen classes.
The other one~\cite{kodirov2015unsupervised,niu2018webly} refines the predicted labels of unseen-class data dynamically as in self-training. 
It is worth noting that this kind of method often shares the same projection in the seen and the unseen class domains, which may be less discriminative since the provided semantic prototypes suffer from the properties of incompleteness and non-visual components.
\textcolor{black}{Unlike existing transductive ZSL models, we creatively formulate a domain adaptation term, which can learn the visual prototypes and super-prototypes of all unseen classes. Specifically, instead of the popular used image-attribute projection, the image-label projection via prototypes can mitigate the domain shift caused by appearance variations of each attribute across all the classes. While the gap of seen- and unseen- class distributions is bridged by sharing interpretable super-prototypes.}

\textbf{Projection Function.}
From the view of constructing the image-semantics interactions, existing ZSL approaches fall into four categories. 
The first one learns a projection function from a visual feature space to a semantic space with a linear~\cite{lampert2014attribute,bansal2018zero,li2018discriminative} or non-linear model~\cite{morgado2017semantically,socher2013zero,yu2018stacked}. The test data are then classified by matching the visual representations in the semantic embedding space with the provided semantic prototypes of unseen classes. 
\textcolor{black}{The second group~\cite{annadani2018preserving,niu2018webly,zhang2017learning,wang2018zero} chooses the reverse projection direction, i.e., from the semantic to visual spaces, to alleviate the hubness problem~\cite{radovanovic2010hubs} caused by nearest neighbour search in a high dimensional space.}
The test data are then classified by resorting to the most similar visual exemplars in the unseen class domain. To capture more distribution information from visual space, recent work focuses on generating pseudo examples for unseen classes with seen-class data~\cite{guo2017zero}, web data~\cite{niu2018webly}, Generative Adversarial Networks~\cite{xian2018feature,zhu2018generative}, Variational Autoencoder ~\cite{schonfeld2019generalized}, etc. Consequently, zero-shot recognition degenerates into a general supervised learning problem.

The third group is a combination of the first two groups but with an additional reconstruction constraint for visual samples or semantic prototypes~\cite{kodirov2017semantic,annadani2018preserving,zhao2018domain}.
Such ZSL approaches generally take the encoder-decoder paradigm, and then conduct the final recognition by the same search strategy as in the first two groups. This makes the projection function generalize better from the seen to the unseen classes as demonstrated in other problems~\cite{ap2014autoencoder}. 
The last group learns a common space, where both the visual space and semantic space are projected to~\cite{liu2018generalized,changpinyo2017predicting,hubert2017learning}. In such a framework, a score function is first trained using seen-class labelled examples, and then computes a likelihood score of the test sample.

Inspired by the third group, we propose to learn interpretable visual prototypes for zero-shot recognition by bidirectional projection. In particular, instead of popularly used image-attribute projection, we adopt image-label projection to avoid those problems caused by the provided semantic prototypes. Additionally, unlike many existing two-step ZSL approaches, the proposed HPL model can perform one-step recognition due to the visual prototype learning.  

\section{Prototype Learning for Zero-shot Recognition}
In this section, we first set up the zero-shot recognition problem~(Section~\ref{Setup}), then develop a HPL model for this task~(Section~\ref{ModelFormulation}), and finally derive an efficient algorithm to solve HPL~(Section~\ref{ModelOptimization}).

\subsection{Problem Definition}\label{Setup}
\begin{table*}[!t]
 \begin{center}
  \caption{Key notations} 
 \label{tab:notation}
   \begin{tabular}{c|l}
    \hline 
     \textbf{Notation}  & \textbf{Description}       \\
     \hline
     $\mathcal{S}=\left \{s_{1},\cdots,s_{m}\right \}$&Set of $m$ seen classes\\ \hline
     $\mathcal{U}=\left \{u_{1},\cdots,u_{n}\right \}$&Set of $n$ unseen classes\\ \hline
     $\bm Y_{s}=\left [ \bm y_{1}^{(s)},\cdots,\bm y_{m}^{(s)}\right]\in \mathbb{R}^{k\times m}$&Set of semantic prototypes of all seen classes\\ \hline
     $\bm Y_{u}=\left [ \bm y_{1}^{(u)},\cdots,\bm y_{n}^{(u)}  \right ]\in \mathbb{R}^{k\times n}$&Set of semantic prototypes of all unseen classes\\
     \hline
     $\mathcal{X},\mathcal{Y}=\left[\bm Y_{s},\bm Y_{u}\right]$&Visual space and semantic space, respectively\\ \hline
     $N_{{s}},N_{{u}}$ & Number of training samples and number of test samples, respectively\\ \hline
     $\left (\bm x_{i}^{(s)} , l_{i}^{(s)},\bm c_{l_{i}^{(s)}}^{(s)}, \bm y_{l_{i}^{(s)}}^{(s)} \right )$ & The $i$-th training sample: image embedding $x_{i}^{{(s)}}\in \mathcal{X} $, and label  $l_{i}^{(s)}\in\mathcal{S}$ with one-hot vector $\bm c_{l_{i}^{(s)}}^{(s)}$ and $\bm y_{l_{i}^{(s)}}^{(s)} \in \bm Y_{s}$ \\ \hline
    $\left ( \bm x_{i}^{(u)} , l_{i}^{(u)},\bm c_{l_{i}^{(u)}}^{(u)}, \bm y_{l_{i}^{(u)}}^{(u)} \right )$& The $i$-th test sample: image embedding $x_{i}^{{(u)}}\in \mathcal{X} $, and label  $l_{i}^{(u)}\in\mathcal{U}$ with one-hot vector $\bm c_{l_{i}^{(u)}}^{(u)}$ and $\bm y_{l_{i}^{(u)}}^{(u)} \in \bm Y_{u}$ \\ \hline
    $k,d$& Dimension of each semantic prototype and dimension of each image embedding, respectively \\ \hline
    $\bm P_{s}=\left [ \bm p_{1}^{(s)},\cdots,\bm p_{m}^{(s)}\right]\in \mathbb{R}^{d\times m}$ &Set of visual prototypes of all seen classes, and $\bm p_{i}^{{(s)}}\in \mathcal{X} $\\ \hline
    $\bm P_{u}=\left [ \bm p_{1}^{(u)},\cdots,\bm p_{n}^{(u)}  \right ]\in \mathbb{R}^{d\times n}$&Set of visual prototypes of all unseen classes, and $\bm p_{i}^{{(u)}}\in \mathcal{X} $\\ \hline 
    $\bm D_{v}=\left [ \bm d_{1}^{(v)},\cdots,\bm d_{q}^{(v)}\right]\in \mathbb{R}^{d\times q}$ &Set of visual super-prototypes, and $\bm d_{i}^{{(v)}}\in \mathcal{X} $\\ \hline
    $\bm D_{c}=\left [ \bm d_{1}^{(c)},\cdots,\bm d_{q}^{(c)}  \right ]\in \mathbb{R}^{k\times q}$&Set of semantic super-prototypes, and $\bm d_{i}^{{(c)}}\in \mathcal{Y} $\\
    \hline
    \end{tabular}
 \end{center}
\end{table*}

 Let $\mathcal{S}$ and $\mathcal{U}$ denote two disjoint sets of seen classes and unseen classes. 
 Accordingly, let $\bm Y_{s}$ and $\bm Y_{u}$ denote the semantic prototypes (e.g. a $k$-dimensional attribute vector or word vector derived from text for each class) of all seen and unseen classes, respectively.
 Meanwhile, suppose we are given a set of labelled training samples $\mathcal{D}_{s} = \left \{ \left (\bm x_{i}^{(s)} , l_{i}^{(s)},\bm c_{l_{i}^{(s)}}^{(s)}, \bm y_{l_{i}^{(s)}}^{(s)} \right ):i=1,\cdots,N_{s} \right \}$, where $\bm x_{i}^{(s)}\in \mathbb{R}^{d}$ is the $d$-dimensional visual embedding of the $i$-th sample in the training set, and its class label $l_{i}^{(s)}$ belongs to the seen classes set $\mathcal{S}$. $\bm c_{l_{i}^{(s)}}^{(s)}\in \mathbb{R}^{m}$ and $\bm y_{l_{i}^{(s)}}^{(s)}\in \mathbb{R}^{k}$ are the one-hot vector and semantic prototype of $\bm x_{i}^{(s)}$, indicating the label $l_{i}^{(s)}$. Let $\bm X_{s}=\left [ \bm x_{1}^{(s)}\cdots,\bm x_{N_{s}}^{(s)}\right]\in \mathbb{R}^{d\times N_{s}}$ and $\bm C_{s}=\left [ \bm c_{l_{1}^{(s)}}^{(s)},\cdots,\bm c_{l_{N_{s}}^{(s)}}^{(s)}\right]\in \mathbb{R}^{m\times N_{s}}$. 
 Similarly, let 
$\mathcal{D}_{u} = \left \{ \left ( \bm x_{i}^{(u)} , l_{i}^{(u)},\bm c_{l_{i}^{(u)}}^{(u)}, \bm y_{l_{i}^{(u)}}^{(u)} \right ):i=1,\cdots,N_{u} \right \}$ denote a set of unlabelled test samples, where $l_{i}^{(u)}\in \mathcal{U}$ is the unknown label of $\bm x_{i}^{(u)}\in \mathbb{R}^{d}$ in the standard ZSL setting. $\bm c_{l_{i}^{(u)}}^{(u)}\in \mathbb{R}^{n}$ and $\bm y_{l_{i}^{(u)}}^{(u)}\in \mathbb{R}^{k}$ are the one-hot vector and semantic prototype of $\bm x_{i}^{(u)}$, corresponding to the class label $l_{i}^{(u)}$. Here, $\bm X_{u}=\left [ \bm x_{1}^{(u)},\cdots,\bm x_{N_{u}}^{(u)}\right]\in \mathbb{R}^{d\times N_{u}}$ and $\bm C_{u}=\left [ \bm c_{l_{1}^{(u)}}^{(u)},\cdots,\bm c_{l_{N_{u}}^{(u)}}^{(u)}\right]\in \mathbb{R}^{n\times N_{u}}$. 
The goal of zero-shot recognition is to predict the labels of test samples in $\mathcal{D}_{u}$ by learning a classifier $f^{u}(\cdot ):\bm X_{u}\rightarrow \mathcal{U}$.
The key notations used throughout this paper are summarized in Table~\ref{tab:notation}.

\subsection{HPL: Formulation}\label{ModelFormulation}

\begin{figure}[!t]
\begin{center}
\centerline{\includegraphics[width=\columnwidth]{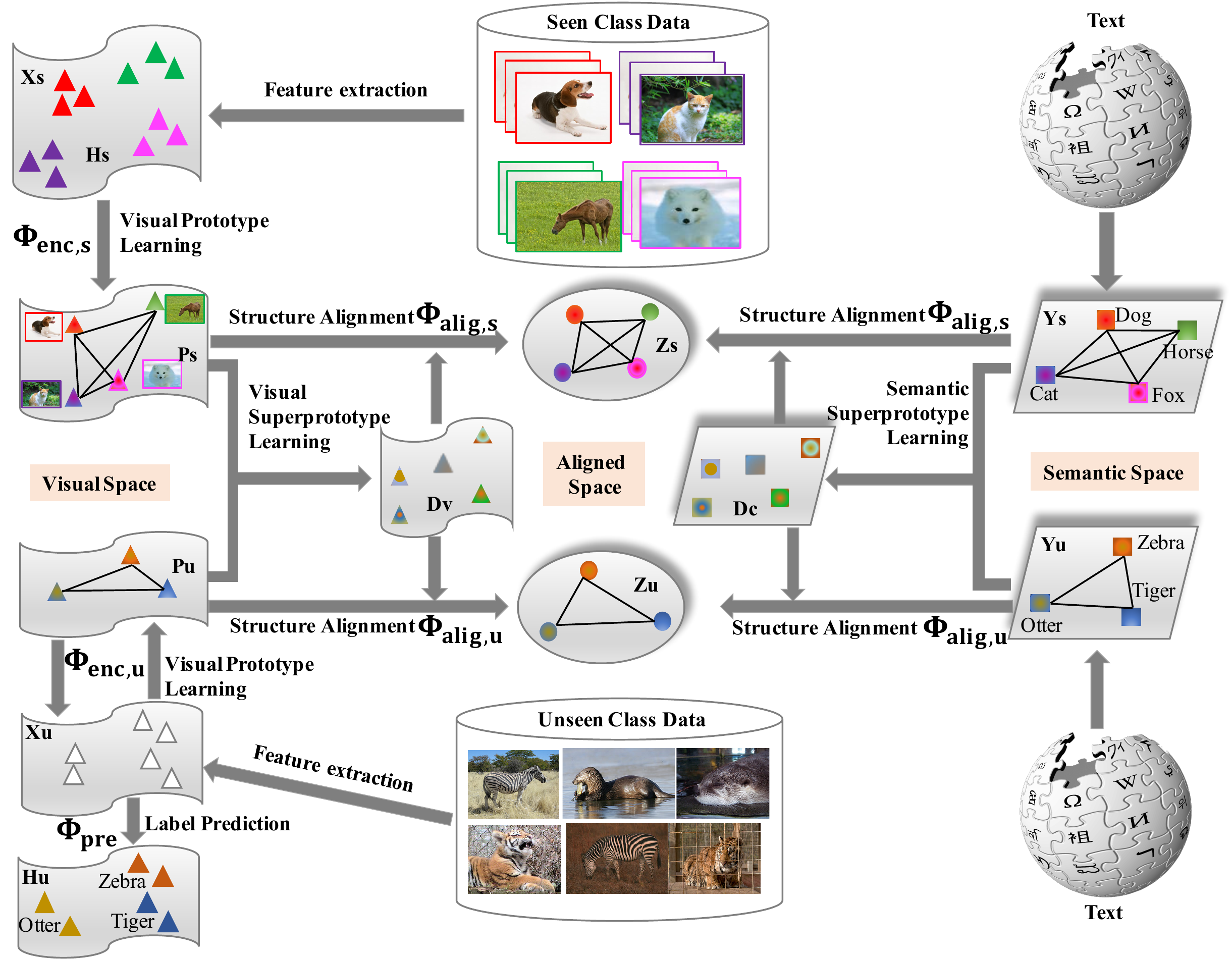}}
\caption{The illustration of HPL model for zero-shot recognition.}
\label{fig:framework}
\end{center}
\vskip -0.2in
\end{figure}

Assume both the seen-class training set $\mathcal{D}_{s}$ and unlabelled unseen-class data $\bm X_{u}$ are available. To predict the labels of $\bm X_{u}$, we propose a hierarchical prototype learning function $\Psi\left ( \bm P_{s},\bm D_{v}, \bm D_{c},\bm Z_{s} \right )$ for zero-shot recognition in an iterative model update process. Specifically, 
$\bm P_{s}=\left [ \bm p_{1}^{(s)},\cdots,\bm p_{m}^{(s)}\right]\in \mathbb{R}^{d\times m}$ denotes the visual prototypes of $\bm X_{s}$, and
$\bm P_{u}=\left [ \bm p_{1}^{(u)},\cdots,\bm p_{n}^{(u)}\right]\in \mathbb{R}^{d\times n}$ denotes the visual prototypes of $\bm X_{u}$.
$\bm D_{v}=\left [ \bm d_{1}^{(v)},\cdots,\bm d_{q}^{(v)}\right]\in \mathbb{R}^{d\times q}$ represents the prototypes of $\bm P_{s}$ and $\bm P_{u}$, named visual super-prototypes, and $\bm D_{c}=\left [ \bm d_{1}^{(c)},\cdots,\bm d_{q}^{(c)}  \right ]\in \mathbb{R}^{k\times q}$ represents the prototypes of $\bm Y_{s}$ and $\bm Y_{u}$, named semantic super-prototypes. Generally, $q \in \{1,\cdots,m+n\}$.
Key to ZSL is to transfer knowledge from the seen to the unseen classes. Motivated by the fact that there exist unseen/seen prototypes that fall into the same class~\cite{zhang2019seeing,zhang2018self}, we thus consider learning super-prototypes to bridge the seen and unseen class domains, and meanwhile, align the visual and semantic spaces. 
$\bm Z_{s}=\left [ \bm z_{1}^{(s)},\cdots,\bm z_{m}^{(s)}\right]\in \mathbb{R}^{q\times m}$ just denotes the structural consistency representations for both two spaces in the seen class domain. 
For discriminative recognition, the minimization of $\Psi$ over all possible assignments, i.e.,
\begin{align} \label{OverPred}
\begin{split}
&\min_{\bm P_{s},\bm D_{v}, \bm D_{c},\bm Z_{s}} \Psi\left ( \bm P_{s},\bm D_{v}, \bm D_{c},\bm Z_{s} \right ) \\
&s.t. \quad \left \| \bm d_{j}^{(v)} \right \|\leq 1, \quad  \left \| \bm d_{j}^{(c)} \right \|\leq 1, \quad  j={1,\cdots,q},
\end{split}
\end{align}
is encouraged to achieve the \underline{three goals} of i) minimizing the encoding cost of $\bm X_{s}$ via visual prototypes $\bm P_{s}$; ii) maximizing the structural consistency between $\bm P_{s}$ and $\bm Y_{s}$ via super-prototypes $\bm D_{v} $ and $\bm D_{c}$ in an aligned space; iii) maximizing the structural consistency between $\bm P_{u}$ and $\bm Y_{u}$ via $\bm D_{v} $ and $\bm D_{c}$ under the constraint of the minimum prediction error of $\bm X_{u}$. It is worth noting that we additionally introduce a regularizer in Eq.~(\ref{OverPred}) for each super-prototype to enhance the stability of solutions and mitigate the scale issue.

For this end, as shown in Fig.~\ref{fig:framework}, we consider a decomposition of the objective function in Eq.~(\ref{OverPred}) into three functions, corresponding to the three aforementioned objectives, as
\begin{align} \label{Obj_decomp}
    \begin{split}
      &  \Psi \left ( \bm P_{s},\bm D_{v}, \bm D_{c},\bm Z_{s} \right )\triangleq \beta \Phi_{\text{enc,s}}\left ( \bm p_{1}^{(s)},\cdots,\bm p_{m}^{(s)} \right )\\
     &   + \Phi_{\text{alig,s}}\left ( \bm P_{s},\bm D_{v},\bm D_{c},\bm Z_{s} \right )   + \gamma \Phi_{\text{pre}}\left ( \bm D_{v}, \bm D_{c} \right )
    \end{split}
\end{align}
where $\Phi_{\text{enc,s}}\left ( \cdot \right )$ is an encoding function that favors learning discriminative prototypes from $\bm X_{s}$ by well encoding $\bm X_{s}$ under the supervision of its labels. $\Phi_{\text{alig,s}}\left ( \cdot \right)$ denotes an alignment function that favors learning interpretable visual and semantic super-prototypes via structure alignment between the visual and semantic spaces. $\Phi_{\text{pre}}\left ( \cdot \right )$ is a prediction function that favors generating more representative super-prototypes with the assistance of predicted labels of $\bm X_{u}$. 
The parameters $\beta,\gamma>0$ control the effects of encoding cost and test data inference on the global objective function $\Psi$. A close to zero $\gamma$ will ignore the prediction error, resulting in poor recognition performance, while a larger $\gamma$ leads to higher recognition accuracy.
Next, we study each of the three functions.

\textbf{Encoding Function:} Inspired by bidirectional projection learning~\cite{kodirov2017semantic}, we adopt both the forward and reverse encoding costs to characterize the discriminability of $\bm P_{s}$ for $\bm X_{s}$. Thus, the encoding function factorizes into
\begin{align}\label{Seen_Prototype}
    \begin{split}
        &\Phi_{\text{enc,s}}\left ( \bm p_{1}^{(s)},\cdots,\bm p_{m}^{(s)} \right )=\Phi_{\text{enc,s}}\left ( \bm P_{s}\right )\\
        &=\sum_{i=1}^{N_{s}} \left (  \left \|\bm P_{s}^{\text{T}} \bm x_{i}^{(s)}- \bm c_{l_{i}^{(s)}}^{(s)}\right \|_{2}^{2}+\left \| \bm x_{i}^{(s)}- \bm P_{s} \bm c_{l_{i}^{(s)}}^{(s)}\right \|_{2}^{2} \right ).
    \end{split}
\end{align}

Unlike the popularly used image-semantics projection in previous works, a feature vector representing the low-level visual appearance of an object is projected into the high-level label space (instead of the middle-level semantic space), and further back to reconstruct itself in our model. In this way, various problems caused by the provided semantic prototypes can be avoided. 
\textcolor{black}{Additionally, the projection in the second term of Eq.~(\ref{Seen_Prototype})\footnote{$\bm x_{i}^{(s)}$ and $\bm p_{j}^{(s)}$ in Eq.~(\ref{Seen_Prototype}) are all normalized vectors for $i=1,\cdots, N_{s}$ and $j=1,\cdots,m$. Then $\bm p_{j}^{(s)^{\text{T}}} \bm x_{i}^{(s)}$ denotes the cosine distance.} encourages the prototype of corresponding class (i.e., $\bm p_{l_{i}^{(s)}}^{(s)}$) is very similar to the sample $\bm x_{i}^{(s)}$, thus guaranteeing the representativeness of learned prototypes. Meanwhile, the reverse projection in the first term enforces $\bm x_{i}^{(s)}$ is closest to its corresponding prototype yet far away from other prototypes, thus learning discriminative prototypes.}

\textbf{Alignment Function:} Since the semantic prototypes are additionally provided for ZSL, we can strengthen the discriminability of visual prototypes by aligning their intrinsic structure with that of semantic prototypes. This is motivated by the fact that there often exist super-prototypes in visual/semantic space, which encourage the original prototypes in the two spaces to be represented consistently in an aligned space, and thus share  the same structure. Therefore, we consider
\begin{align}\label{Super_Prototype}
    \begin{split}
        &\Phi_{\text{alig,s}}\left ( \bm P_{s},\bm D_{v},\bm D_{c},\bm Z_{s} \right )=\\
        &\sum_{i=1}^{m}\left ( \left \| \bm p_{i}^{(s)}-\bm D_{v} \bm z_{i}^{(s)}\right \|_{2}^{2}+\lambda \left \| \bm y_{i}^{(s)}-\bm D_{c} \bm z_{i}^{(s)}\right \|_{2}^{2} \right ),
    \end{split}
\end{align}
where $\lambda$ is a nonnegative parameter controlling the relative importance of the visual and semantic spaces. Additionally, the alignment function enforces the same number of super-prototypes in the two spaces, because the intrinsic structure is unique either in low-level or high-level spaces. In particular, such a structure alignment strategy in Eq.~(\ref{Super_Prototype}) alleviates the indiscriminability issue of prototypes that are learned from Eq.~(\ref{Seen_Prototype}) in the unbalanced data scenario.

\textbf{Prediction Function:} Notice that the super-prototypes learned in the seen class domain are expected to be shared with unseen class domain. Consequently, the unseen-class samples become seen at the superhigh-level semantic space, and thus easier to be recognized~\footnote{For instance, ``dolphin'' belongs to an unseen class, but becomes seen in terms of mammal (a super-class).}. For this end, the prediction function mainly pursues the minimum encoding cost of test samples and the maximum structural consistency of prototypes in the unseen class domain. Thus, we formulate the prediction function as 
\begin{align}\label{Unseen}
    \begin{split}
     &\Phi_{\text{pre}}\left ( \bm D_{v}, \bm D_{c} \right )=  \\
     &\min_{\bm P_{u},\bm C_{u} \subseteq \mathcal{H},\bm Z_{u}} \left ( \beta \Phi_{\text{enc,u}}\left ( \bm P_{u},\bm C_{u} \right ) + \Phi_{\text{alig,u}}\left ( \bm P_{u},\bm Z_{u} \right ) \right ),
    \end{split}
\end{align}
where $\bm Z_{u}=\left [ \bm z_{1}^{(u)},\cdots,\bm z_{n}^{(u)}\right]\in \mathbb{R}^{q\times n}$ denotes the structural consistency representations for both visual and semantic prototypes in the unseen class domain,
\begin{align}\label{Unseen_enc}
    \begin{split}
     &\Phi_{\text{enc,u}}\left ( \bm P_{u},\bm C_{u}\right )=\\
     &\sum_{i=1}^{N_{u}} \left (  \left \|\bm P_{u}^{\text{T}} \bm x_{i}^{(u)}- \bm c_{l_{i}^{(u)}}^{(u)}\right \|_{2}^{2}+\left \| \bm x_{i}^{(u)}- \bm P_{u} \bm c_{l_{i}^{(u)}}^{(u)}\right \|_{2}^{2} \right ),
    \end{split}
\end{align}
\begin{align}\label{Unseen_alig}
    \begin{split}
    &\Phi_{\text{alig,u}}\left ( \bm P_{u},\bm Z_{u} \right )=\\
    &\sum_{i=1}^{n}\left ( \left \| \bm p_{i}^{(u)}-\bm D_{v} \bm z_{i}^{(u)}\right \|_{2}^{2}+\lambda \left \| \bm y_{i}^{(u)}-\bm D_{c} \bm z_{i}^{(u)}\right \|_{2}^{2} \right ),
    \end{split}
\end{align}
and $\mathcal{H}=\{\bm 1^{\text{T}}\bm c_{l_{i}^{(u)}}^{(u)}=1, \bm c_{l_{i}^{(u)}}^{(u)} \in \left \{ 0,1 \right \}^{n}, i={1,\cdots,N_{u}}\}$.

As a result, our HPL model in Eq.~(\ref{OverPred}) casts ZSL into a min-min optimization problem. Different from most existing ZSL approaches that perform the final recognition via nearest neighbor search, the class label of each test sample is predicted directly via the prediction function in our model. Such a one-step recognition framework is also generic in the sense that it can be easily extended to inductive settings by reformulating the prediction function $\Phi_{\text{pre}}^{\text{I}}\left ( \bm D_{v}, \bm D_{c} \right )$ as 
\begin{align}\label{Unseen_indu}
    \begin{split}
     &\Phi_{\text{pre}}^{\text{I}}\left ( \bm D_{v}, \bm D_{c} \right )=  \min_{\bm P_{u},\bm Z_{u}} \Phi_{\text{alig,u}}\left ( \bm P_{u},\bm Z_{u} \right ) .
    \end{split}
\end{align}

Besides the standard ZSL above, generalized zero-shot learning (GZSL) where prediction on test data is made over both seen and unseen classes, has drawn much attention recently~\cite{hubert2017learning,liu2018generalized}. To further improve the generalization ability of our HPL model on GZSL tasks, we reformulate the prediction function $\Phi_{\text{pre}}^{\text{G}}\left ( \bm P_{s},\bm D_{v}, \bm D_{c} \right )$ as  
\begin{align}\label{GZSL}
     &\Phi_{\text{pre}}^{\text{G}}\left ( \bm P_{s},\bm D_{v}, \bm D_{c} \right )=  \\
     &\min_{\bm P_{u},\bm C_{u} \subseteq \mathcal{H}^{\text{G}},\bm Z_{u}} \left ( \beta \Phi_{\text{enc,u}}^{\text{G}}\left ( \bm P_{u},\bm C_{u} \right ) + \Phi_{\text{alig,u}}\left ( \bm P_{u},\bm Z_{u} \right ) \right ),\nonumber
\end{align}
where 
\begin{align}\label{Unseen_enc_GZSL}
     &\Phi_{\text{enc,u}}^{\text{G}}\left ( \bm P_{u},\bm C_{u}\right )=\sum_{i=1}^{N_{u}} \left (  \left \|\left [ \bm P_{s}, \bm P_{u} \right ]^{\text{T}} \bm x_{i}^{(u)}- \bm c_{l_{i}^{(u)}}^{(u)}\right \|_{2}^{2}\right )\nonumber\\
     &+\sum_{i=1}^{N_{u}} \left (\left \| \bm x_{i}^{(u)}- \left [ \bm P_{s}, \bm P_{u} \right ] \bm c_{l_{i}^{(u)}}^{(u)}\right \|_{2}^{2} \right ) ,
\end{align}
and 
$\mathcal{H}^{\text{G}}=\{\bm 1^{\text{T}}\bm c_{l_{i}^{(u)}}^{(u)}=1, \bm c_{l_{i}^{(u)}}^{(u)} \in \left \{ 0,1 \right \}^{m+n}, i={1,\cdots,N_{u}}\}$.
The main difference between Eq.~(\ref{Unseen}) and Eq.~(\ref{GZSL}) lies in encoding $\bm X_{u}$ via both $\bm P_{s}$ and $\bm P_{u}$ under the GZSL setting.

\subsection{HPL: Algorithm}\label{ModelOptimization}
Here, we consider the model optimization under the standard ZSL setting~\footnote{Our proposed optimization can also be generalized to the GZSL setting as presented in the supplementary material.}. Putting all three functions together, we consider the following minimization problem
\begin{align} \label{OverPred_v1}
&\min_{\bm P_{s},\bm D_{v}, \bm D_{c},\bm Z_{s}} \Big\{\beta \Phi_{\text{enc,s}}\left ( \bm P_{s} \right )+ \Phi_{\text{alig,s}}\left ( \bm P_{s},\bm D_{v},\bm D_{c},\bm Z_{s} \right )  \nonumber \\
&  + \gamma \min_{\bm P_{u},\bm C_{u} \subseteq \mathcal{H},\bm Z_{u}} \left ( \beta \Phi_{\text{enc,u}}\left ( \bm P_{u},\bm C_{u} \right ) + \Phi_{\text{alig,u}}\left ( \bm P_{u},\bm Z_{u} \right ) \right )\Big\} \nonumber\\
&s.t. \quad \left \| \bm d_{j}^{(v)} \right \|\leq 1, \quad  \left \| \bm d_{j}^{(c)} \right \|\leq 1, \quad  j={1,\cdots,q}.
\end{align}

It is not trivial to solve the optimization problem in Eq.~(\ref{OverPred_v1}), since the last term in the objective function is also a minimization problem. In the following, we will formulate our solver as an iterative optimization algorithm. 
Given the two super-prototype sets $\bm D_{v}^{(t)}$ and $\bm D_{c}^{(t)}$ at iteration $t$ during model learning, we can obtain the optimal solution $\{\bm P_{u}^{(t+1)},\bm C_{u}^{(t+1)},\bm Z_{u}^{(t+1)}\}$ by solving the optimization problem in Eq.~(\ref{Unseen}). Then, the optimization problem in Eq.~(\ref{OverPred_v1}) at iteration $t+1$ can be approximated as follows
\begin{align} \label{ObjectApproximate1}
\begin{split}
&\min_{\bm P_{s}^{(t+1)},\bm D_{v}^{(t+1)}, \bm D_{c}^{(t+1)},\bm Z_{s}^{(t+1)}}\quad
 \beta \Phi_{\text{enc,s}}\big (\bm P_{s}^{(t+1)} \big )+\\
 &\Phi_{\text{alig,s}} \big( \bm P_{s}^{(t+1)},\bm D_{v}^{(t+1)}, \bm D_{c}^{(t+1)},\bm Z_{s}^{(t+1)} \big ) +\\
 &\gamma  \Phi_{\text{pre}}\left ( \bm D_{v}^{(t+1)}, \bm D_{c}^{(t+1)} \right ) \\
&s.t. ~  \left \| \bm d_{j}^{(v)^{(t+1)}} \right \|\leq 1,   \left \| \bm d_{j}^{(c)^{(t+1)}} \right \|\leq 1,   j={1,\cdots,q},
\end{split}
\end{align}
where $\Phi_{\text{pre}}\left ( \bm D_{v}^{(t+1)}, \bm D_{c}^{(t+1)} \right )=\beta \Phi_{\text{enc,u}}\left ( \bm P_{u}^{(t+1)},\bm C_{u}^{(t+1)} \right ) + \Phi_{\text{alig,u}}\left ( \bm P_{u}^{(t+1)},\bm Z_{u}^{(t+1)} \right ) $.

As summarized in Algorithm~\ref{alg:Framwork}, our solver consists of iterating between \emph{updating unseen-class data prediction} and \emph{updating seen-class data fitting}.
It is obvious that the optimization problem in Eq.~(\ref{Unseen}) (resp. Eq.~(\ref{ObjectApproximate1})) is not convex for the three (resp. four) variables simultaneously, but it is convex for each of them separately. We thus employ an alternative optimization method to solve it. The details about solving Eq.~(\ref{Unseen}) and Eq.~(\ref{ObjectApproximate1}) are provided in the supplementary material.
Particularly, the objective function in Eq.~(\ref{ObjectApproximate1}) can be further simplified equivalently as
\begin{align}\label{SeenOpt1}
\begin{split}
& \mathcal{L}(\bm P_{s}^{(t+1)},\bm D_{v}^{(t+1)}, \bm D_{c}^{(t+1)},\bm Z_{s}^{(t+1)})\\
&=\rho(1-\omega)(1-\alpha)\Phi_{\text{enc,s}}\big (\bm P_{s}^{(t+1)} \big ) \\
& +(1-\rho)(1-\omega)(1-\alpha) \left \| \bm P_{s}^{(t+1)}-\bm D_{v}^{(t+1)}\bm Z_{s}^{(t+1)}\right \|_{F}^{2} \\
&+ \omega (1-\rho)(1-\alpha) \left \| \bm Y_{s}-\bm D_{s}^{(t+1)}\bm Z_{s}^{(t+1)} \right \|_{F}^{2}  \\
&  + \alpha (1-\rho)(1-\omega) \left \| \bm P_{u}^{(t+1)} -\bm D_{v}^{(t+1)}\bm Z_{u}^{(t+1)} \right \|_{F}^{2} \\
&+\omega \alpha (1-\rho) \left \| \bm Y_{u}-\bm D_{c}^{(t+1)}\bm Z_{u}^{(t+1)}  \right \|_{F}^{2}  ,
\end{split}
\end{align}
where $\rho = \beta/(1+\beta)$, $\omega =\lambda/(1+\lambda)$, $\alpha =\gamma/(1+\gamma)$, and thus $\rho,\omega,\alpha \in \left [ 0,1 \right)$. This can facilitate the parameters tuning. 

\begin{algorithm}[tb]
  \caption{HPL for Zero-Shot Recognition}
  \label{alg:Framwork}
\begin{algorithmic}[1]
  \STATE {\bfseries Input:} 
  $\mathcal{D}_{s}$ (training set); $\bm X_{u}$ (test samples); $\bm Y_{u}$ (semantic prototypes of unseen classes); parameters ($\rho,\omega,\alpha$); $q$ (the number of super-prototypes); ($\varepsilon$,~$maxIter$).
  \STATE Initialize $t\leftarrow 0$; $\bm D_{v}^{(t)}$, $\bm D_{c}^{(t)}$.
  \STATE {\bfseries Output:} $\bm C_{u}^{*}$.
  \REPEAT
  \STATE Update \{$\{\bm P_{u}^{(t+1)},\bm C_{u}^{(t+1)},\bm Z_{u}^{(t+1)}\}\}$ via Eq.~(\ref{Unseen});
  \STATE Update $\{\bm P_{s}^{(t+1)},\bm D_{v}^{(t+1)}, \bm D_{c}^{(t+1)},\bm Z_{s}^{(t+1)}\}$ via Eq.~(\ref{ObjectApproximate1});
  \STATE $Err_{1}=\left \| \bm D_{v}^{(t+1)}-\bm D_{v}^{(t)} \right \|_{2}$;
  \STATE $Err_{2}=\left \| \bm D_{s}^{(t+1)}-\bm D_{s}^{(t)} \right \|_{2}$;
  \STATE $t\leftarrow t+1$;
  \UNTIL{($Err_{1}<\varepsilon$ and $Err_{2}<\varepsilon$) or ($t>maxIter$)}
  \STATE $\bm D_{v} \leftarrow \bm D_{v}^{(t)}$, $\bm D_{c} \leftarrow \bm D_{c}^{(t)}$;
  \STATE \color{black}{Obtain $\bm C_{u}^{*}$, $\bm P_{u}^{*}$, and $\bm Z_{u}^{*}$ via Eq.~(\ref{Unseen}).}
\end{algorithmic}
\end{algorithm}

\textbf{Convergence Analysis.}
As can be observed from Eq.~(\ref{Unseen}) and Eq.~(\ref{ObjectApproximate1}), due to linear formulations, it is easier to solve the seven sub-problems corresponding to the seven variables in our model. Specifically, the solutions to $\bm Z_{s}$ and $\bm Z_{u}$ can be expressed in closed forms, updating $\bm P_{s}$ and $\bm P_{u}$ is actually to solve Sylvester equations~\cite{bartels1972solution}, and computing $\bm C_{u}$ is to perform minimum search. Moreover, we adopt a line search strategy~\cite{more1994line} when updating $\bm D_{v}$ and $\bm D_{c}$. Thus, the objective function in Eq.~(\ref{OverPred_v1}) is non-increasing with a lower bound during the iterative optimization of each sub-problem as in Algorithm~\ref{alg:Framwork}.

\textbf{Complexity Analysis.}
We further analyze the time complexity for Algorithm~\ref{alg:Framwork} as follows. The complexity of solving Eq.~(\ref{Unseen}) is $\mathcal{O}(\nu_{1}(n^{2}dN_{u}+d^{3}+(q+d+k)q^{2}))$, and updating \{$\bm D_{v}^{(t+1)}$, $\bm D_{c}^{(t+1)}$\} by solving Eq.~(\ref{ObjectApproximate1}) spends $\mathcal{O}(\nu _{2}(d^{3}+m^{3}+(d+k+m+n)q^{2}+(m+n)(d+k)q))$~\footnote{The computation complexity about training samples $\bm X_{s}$ and $\bm C_{s}$ is excluded due to storage in advance.}, where $\nu_{1}$ and $\nu_{2}$ are the numbers of (inner) iterations required to converge. To sum up, one iteration in Algorithm~\ref{alg:Framwork} has a linear time complexity of $\mathcal{O}(\nu_{1}n^{2}dN_{u}+(\nu_{1}+\nu_{2})(d^{3}+(m+n+d)q^{2}))$ ($d,q,m,n\ll N_{u}$) with respect to the test data size $N_{u}$. Thus, it is efficient even for large-scale ZSL problems.

\section{Experimental Results and Analysis}
In this section, we first detail our experimental protocol, and then present the experimental results by comparing our HPL model with the state of the art for zero-shot recognition on five benchmark datasets under various settings.

\begin{table*}[t]
\caption{Statistics for five datasets. $m$ consists of training and validation classes, and $N_{u}$ under the PS protocol includes seen- and unseen- class samples. Notice that SS and PS protocols have no effects on ImageNet dataset.} 
\label{tab:datasets}
\begin{center}
\begin{small}
\begin{sc}
\begin{tabular}{cc|ccc|ccccc}
    \hline \hline
    \multicolumn{5}{c}{} & \multicolumn{5}{|c}{\textbf{Number of Images}}\\
    \multicolumn{2}{c}{} & \multicolumn{3}{c|}{\textbf{Number of Classes}} & & \multicolumn{2}{c}{\textbf{SS}}& \multicolumn{2}{c}{\textbf{PS}}\\
     \textbf{Dataset} & $k$ & \textbf{Total} & $m$ & $n$ & \textbf{Total} & $N_{s}$ & $N_{u}$ & $N_{s}$ & $N_{u}$ \\
     aPY & 64 & 32 & 15+5 & 12 & 15339 & 12695 & 2644 &5932 & 1483+7924\\
     AwA  & 85 & 50 & 27+13 & 10 & 30475 & 24295 & 6180 & 19832 & 4958+5685 \\
     SUN & 102 & 717 & 580+65 & 72 & 14340 & 12900 & 1440 & 10320 & 2580+1440\\
     CUB & 312 & 200 & 100+50 & 50 & 11788 & 8855 & 2933 & 7057 &1764+2967 \\
     ImageNet & 1000 & 1360 & 800+200 & 360 & 254000 & 200000 & 54000 & - & -\\
    \hline\hline
    \end{tabular}
\end{sc}
\end{small}
\end{center}
\vskip -0.2in
\end{table*}

\subsection{Evaluation Setup and Metrics}
\textbf{Datasets.} Among the most widely used datasets for ZSL, we first select four attribute datasets. Two of them are coarse-grained, one small (aPascal \& Yahoo (aPY)~\cite{farhadi2009describing}) and one medium-scale (Animals with Attributes (AwA)~\cite{lampert2014attribute}). Another two datasets (SUN Attribute (SUN)~\cite{patterson2012sun} and CUB-200-2011 Birds (CUB)~\cite{wah2011caltech}) are both fine-grained and medium-scale. We also additionally adopt a large-scale dataset (ImageNet~\cite{russakovsky2015imagenet}) for standard ZSL, where 1K classes of ILSRVC 2012 are used as seen classes, and non-overlapping 360 classes from the ILSVRC 2010 serve as unseen classes as in~\cite{kodirov2017semantic}. For ImageNet, we use Word2Vec~\cite{mikolov2013distributed} trained on Wikipedia provided by~\cite{changpinyo2016synthesized} since attributes of 21K classes are not available. Details of all dataset statistics are in Table~\ref{tab:datasets}.

\textbf{Protocols.} 
For fair comparisons, we conduct extensive experiments based on two typical protocols as shown in Table~\ref{tab:datasets}: Standard Split protocol (SS)~\cite{lampert2014attribute} and Proposed Split protocol (PS)~\cite{xian2018zero}. The main difference between SS and PS is that PS can guarantee no unseen classes are from ImageNet-1K since it is used to pre-train the base network, otherwise the zero-shot rule would be violated. Specifically, we will conduct standard ZSL task under both SS and PS protocols, while conduct GZSL task just under the PS protocol.

\textbf{Visual Features.} 
Due to different visual features used by existing ZSL approaches under the SS protocol, we choose to compare with them based on three types of widely-used features: 1024-dim GoogLeNet features (G), 2048-dim ResNet-101 features (R), and 4096-dim VGG19 features (V) provided by~\cite{changpinyo2016synthesized},~\cite{xian2018zero}, and~\cite{zhang2015zero}, respectively. This enables a direct comparison with the published results of existing methods. While under the PS protocol, all compared methods are based on ResNet-101 features since they usually yield higher accuracy than other features as demonstrated in~\cite{xian2018zero}.

\textbf{Evaluation Metrics.} 
\textcolor{black}{At test phase of ZSL, we are interested in having high performance on both densely and sparsely populated classes. Thus, we use the unified evaluation protocol proposed in~\cite{xian2018zero}, where the average accuracy is computed independently for each class. Specifically, under the standard ZSL setting, we measure average per-class top-1 accuracy by}
\begin{align*}
Acc_{\mathcal{U}}=\frac{1}{n}\sum_{i=1}^{n}\frac{\text{\#correct predictions in}~u_{i}}{\text{\# samples in}~u_{i}}.
\end{align*}

In particular, the average per-class top-5 accuracy is computed for ImageNet dataset.
While under the GZSL setting, we compute the harmonic mean ($H$) of $Acc_{\mathcal{S}}$ and $Acc_{\mathcal{U}}$ to favor high accuracy on both seen and unseen classes:
\begin{align*}
H=\frac{2*Acc_{\mathcal{S}}*Acc_{\mathcal{U}}}{Acc_{\mathcal{S}}+Acc_{\mathcal{U}}},
\end{align*}
where $Acc_{\mathcal{S}}$ and $Acc_{\mathcal{U}}$ are the accuracy of recognizing the test samples from the seen and unseen classes respectively, and 
\begin{align*}
Acc_{\mathcal{S}}=\frac{1}{m}\sum_{i=1}^{m}\frac{\text{\#correct predictions in}~s_{i}}{\text{\# samples in}~s_{i}}.
\end{align*}

\textbf{Parameter Settings.} 
There are four parameters in our HPL model: $\rho,\omega,\alpha \in \left [ 0,1 \right)$, and $q \in \{1,\cdots,m+n\}$ (the number of super-prototypes). As in~\cite{zhang2016zero,kodirov2017semantic,xian2018feature}, these hyperparameters are also fine-tuned on a disjoint set of validation set of $5,13,65,50,200$ classes for each dataset, respectively.

\begin{figure}[t!]
\subfigure[The convergence results]{\label{fig:conv}
\begin{minipage}[t]{0.4\linewidth}
\centering
\includegraphics[width=1.5in]{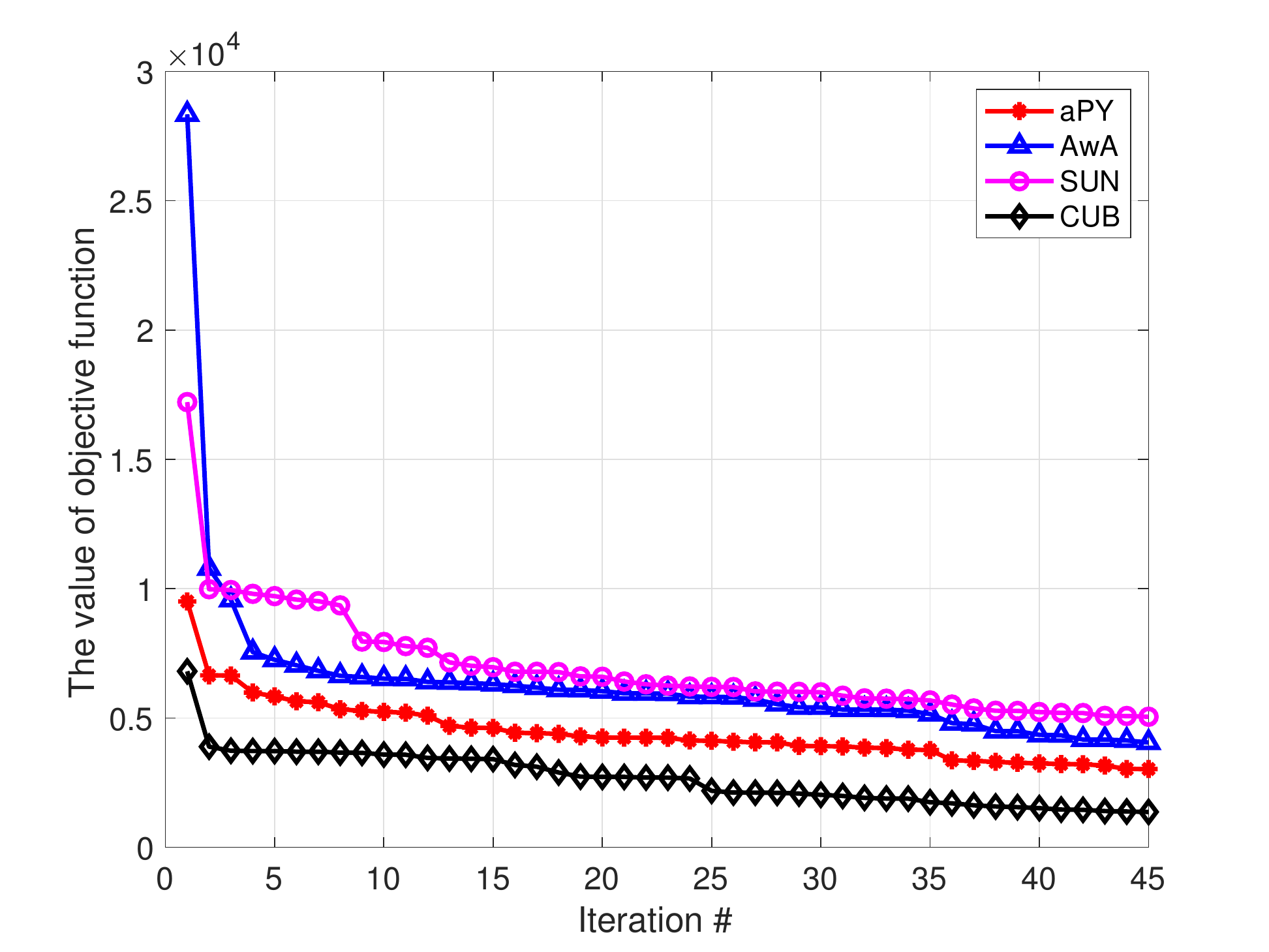}
\end{minipage}%
}
\subfigure[The complexity results]{\label{fig:time}
\begin{minipage}[t]{0.4\linewidth}
\centering
\includegraphics[width=1.5in]{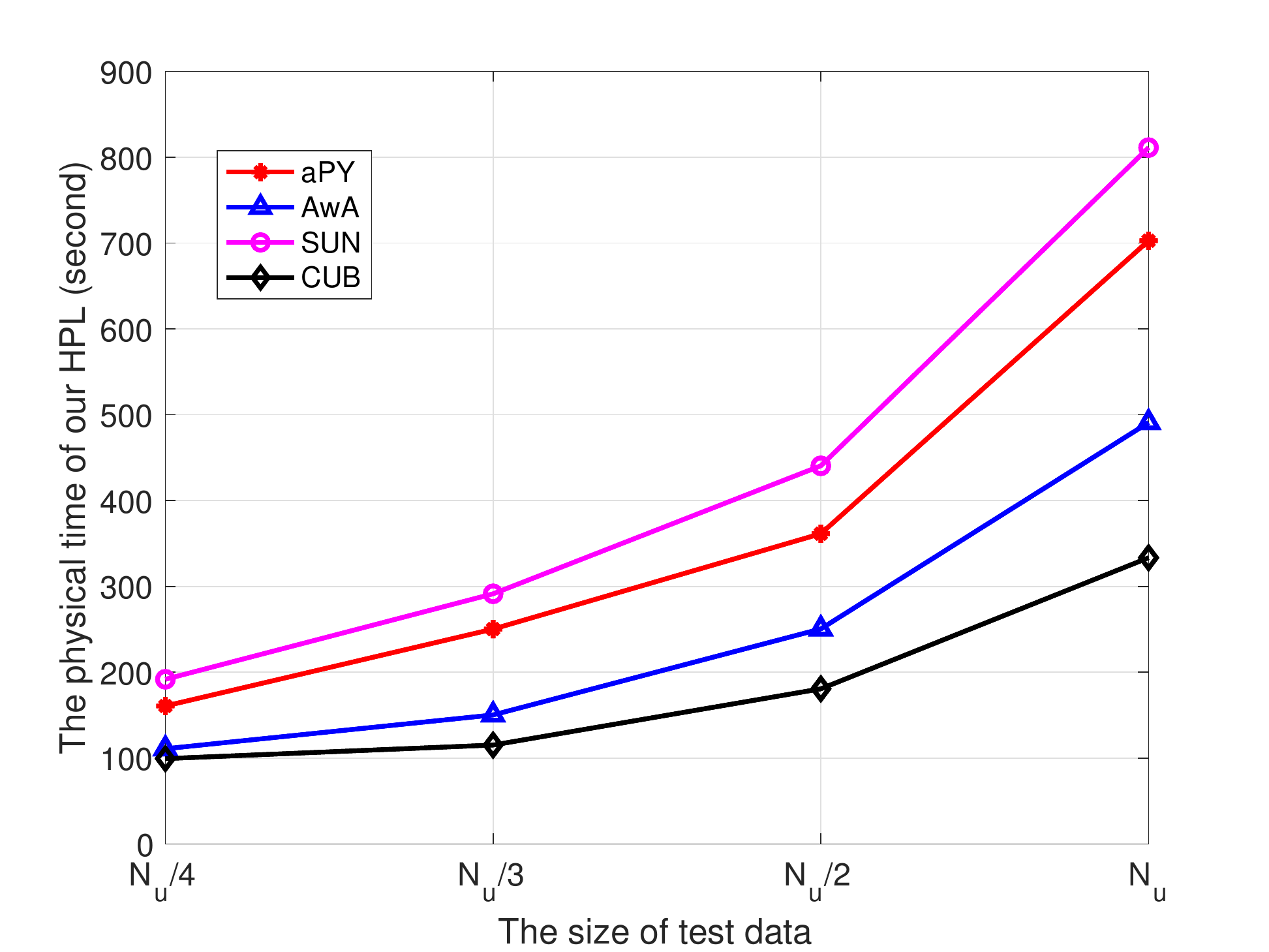}
\end{minipage}%
}
\centering
\caption{The empirical results about convergence and complexity performances of our algorithm for standard ZSL on four datasets under the PS protocol.}
\label{fig:conv_time}
\vspace{-0.15in}
\end{figure} 

\textbf{Compared Methods.} We choose to compare with a wide range of competitive and representative ZSL approaches, especially those that have achieved the state-of-the-art results recently. In particular, such compared approaches involve not only both inductive and transductive models, but also both shallow and deep models.

\begin{table*}[t]
\caption{Comparative results (\%) of standard ZSL on five datasets under the SS protocol. Notations - 'Ind.': Inductive; 'Trans.': Transductive; 'S': Shallow; 'D': Deep; '-': No result reported in the original paper. For each dataset, the best result is marked in \textbf{bold} font and the second best is in blue. We report results averaged over 6 random trails. 
 } 
\label{tab:SS_SZSL}
\begin{center}
\begin{small}
\begin{sc}
\begin{tabular}{c|c|c|c|ccccc}
    \hline \hline
    \textbf{Type} & \textbf{Method} & \textbf{Fea} & \textbf{Model} & \textbf{aPY} & \textbf{AwA} & \textbf{SUN} & \textbf{CUB} & \textbf{ImageNet} \\
    \hline
     \multirow{9}*{Ind.} & ESZSL~\cite{romera2015embarrassingly} & R & S & 34.4 & 74.7 & 57.3 & 55.1 & -\\
    & SynC~\cite{changpinyo2016synthesized}& G & S & - & 72.9 & 62.7 & 54.7 & -\\
    & SAE~\cite{kodirov2017semantic} & G & S &55.4 & 84.7 & 65.2 & 61.4 & 27.2 \\
    & EXEM~\cite{changpinyo2017predicting} & G & S & - &77.2 & 69.6 & 59.8 &-\\
    & GANZrl~\cite{tong2018adversarial} &G&D&-&-&-&62.6 &29.6\\
    & CAPD~\cite{rahman2018unified} & G &S & 55.1& 80.8& - &45.3&23.6\\
    & DCN~\cite{liu2018generalized} & G & D & - &82.3 & 67.4 &55.6 &-\\
    & SE-ZSL~\cite{verma2018generalized}& R & D & - & 83.8 & 64.5 &60.3 &25.4\\
    & MSplit LBI~\cite{zhao2018msplit} & V & S& -&85.3 & - &57.5 & 18.8\\
    \hline
    \multirow{13}*{Trans.}& TMV-HLP~\cite{fu2015transductive}& V & S&-&80.5&-&47.9&-\\
    &SP-ZSR~\cite{zhang2016zero}& V & S & 69.7 & 92.1 & - & 55.3 & -\\
    &GFZSL~\cite{verma2017simple}&V & S &-&94.3 & $\bm{87.0}$ &63.7 &-\\
    &DSRL~\cite{ye2017zero}&V & S & 56.3 & 87.2 & 85.4 &  57.1 &-\\
    & BiDiLEL~\cite{wang2017zero} &G & S  &- & 92.6& - & 62.8 & - \\
    & STZSL~\cite{guo2017zero} & V & S & 54.4 & 83.7 & - & 58.7 & -\\
    & TSTD~\cite{yu2018transductive} & V & S & - & 90.3 &- & 58.2 &-\\
    & DIPL~\cite{zhao2018domain} & G & S &87.8 & 96.1 & 70.0 & 68.2 & \textcolor{blue}{31.7} \\
    & VZSL~\cite{wang2018zero}&V &D & - & 94.8 &- & 66.5 & 23.1\\
    & QFSL~\cite{song2018transductive} & G & D & -&-& 61.7 &69.7 & -\\
    \cline{2-9}
    & HPL & G & S &89.2  & \textcolor{blue}{96.3} & \textcolor{blue}{85.8} & \textcolor{blue}{72.1} & 29.2 \\
    & HPL & R & S &$\bm{91.1}$ & $\bm{97.8}$ & 80.4 & $\bm{75.3}$ & 27.3 \\
    & HPL & V & S &\textcolor{blue}{89.7} & 95.5 & 81.5 & 70.8 &  $\bm{32.6}$\\
    \hline\hline
    \end{tabular}
\end{sc}
\end{small}
\end{center}
\end{table*}

\begin{table*}[t]
\caption{Comparative results (\%) of standard ZSL on four datasets with ResNet-101 features under the PS protocol.} 
\label{tab:PS_SZSL}
\begin{center}
\begin{small}
\begin{sc}
\begin{tabular}{c|c|c|cccc}
    \hline \hline
    \textbf{Type} & \textbf{Method} & \textbf{Model} & \textbf{aPY} & \textbf{AwA} & \textbf{SUN} & \textbf{CUB}  \\
    \hline
     \multirow{9}*{Ind.} 
     & ESZSL~\cite{romera2015embarrassingly} & S & 38.3 & 58.2 & 54.5 & 53.9\\
    & SynC~\cite{changpinyo2016synthesized}& S & 23.9 & 54.0 & 56.3 & 55.6 \\
     & SAE~\cite{kodirov2017semantic} & S & 8.3 & 53.0 & 40.3 & 33.3\\ 
    & CAPD~\cite{rahman2018unified} & S & 39.3 & 52.6 & 49.7 & 53.8 \\
    & f-CLSWGAN~\cite{xian2018feature} & D & - &69.9 & 62.1 & 61.5  \\
    & CDL~\cite{jiang2018learning}  & S &43.0 & 69.9 & 63.6 & 54.5   \\
    & DCN~\cite{liu2018generalized}  & D & 43.6 &65.2 & 61.8 & 56.2 \\
    & SE-ZSL~\cite{verma2018generalized} & D & - & 69.5 &63.4 &59.6 \\
    & PreseR~\cite{annadani2018preserving} & D & 38.4 & - &61.4 &56.0 \\
    \hline
    \multirow{6}*{Trans.}
    &ALE~\cite{akata2016label}&S&45.5&65.3&56.1&54.3\\
    &GFZSL~\cite{verma2017simple}&S&36.9&81.5&63.5&50.4\\
    &DSRL~\cite{ye2017zero}&S&44.8&74.1&57.2 & 48.9\\
    & DIPL~\cite{zhao2018domain}  & S &\textcolor{blue}{69.6} & \textcolor{blue}{85.6 }& \textcolor{blue}{67.9} & 65.4 \\
    & QFSL~\cite{song2018transductive} & D & -&-& 58.3 &\textcolor{blue}{72.1} \\
    \cline{2-7}
    & HPL & S &$\bm{73.8}$ & $\bm{91.2}$ & $\bm{70.4}$ & $\bm{75.2}$ \\
    \hline\hline
    \end{tabular}
\end{sc}
\end{small}
\end{center}
\vskip -0.2in
\end{table*}

\subsection{Comparative Results}
\textbf{Standard ZSL.}
We firstly compare our HPL model with existing state-of-the-art ZSL approaches under the standard setting. Experiments are conducted on five datasets. We use both the SS and PS protocols for more convincing results. To further verify that our method is not only effective to specific visual features, we implement our model under the SS protocol with 1024-dim GoogLeNet features (G), 2048-dim ResNet-101 features (R), and 4096-dim VGG19 features (V) separately. 
The comparative results under the SS and PS protocols are reported in Table~\ref{tab:SS_SZSL} and Table~\ref{tab:PS_SZSL}, respectively. 

It can be seen that, 1) our HPL model yields better performance than the state-of-the-art baselines. This validates that by minimizing encoding cost and maximizing structural consistency, the learned prototypes are discriminative enough to recognize the unseen-class samples.
2) For our model, ResNet and VGG19 features generally lead to better results than GoogLeNet features, except on the SUN dataset. This is due to the fact that only scarce (about 20) training samples are available for each class in the SUN dataset, thus resulting in over-fitted models (e.g., ResNet101).
3) For five datasets, the improvements obtained by our model over the strongest competitor range from 0.9\% to 5.6\%. This actually \underline{creates new baselines} in the area of ZSL, given that most of the compared models take far more complicated nonlinear formulations or even generate numerous training samples for unseen classes. 4) With the assistance of test samples, our model performs better than those inductive approaches under either SS or PS protocol. However, by comparing Table~\ref{tab:PS_SZSL} with Table~\ref{tab:SS_SZSL}, almost all ZSL approaches suffer from performance degradation under the PS protocol, which comes from the fact that the unseen class information is removed from the dataset that is used to pre-train the base network for feature extraction.
5) Our model outperforms those image-attribute projection based approaches obviously (e.g., SAE~\cite{kodirov2017semantic} and QFSL~\cite{song2018transductive}), which demonstrates the effectiveness of prototype learning via image-label projection. 

Additionally, in our experiments, we set the maximum iterations as 100 and the optimization always converges after tens of iterations, usually less than 60. 
\textcolor{black}{As shown in Fig.~\ref{fig:conv}, the objective function of our HPL model is obviously non-increasing and finally converges with the proposed iterative update algorithm. Meanwhile, we also report the physical running time of our algorithm on four datasets under PS protocol for standard ZSL task in Fig.~\ref{fig:time}, which indicates that our iterative update algorithm has linear time complexity with respect to the test data size $N_{u}$. These observations finally support the theoretical analysis of convergence and complexity in Section~\ref{ModelOptimization}. Thus, the proposed algorithm is indeed efficient especially compared with those approaches taking far more complicated nonlinear formulations.}


\begin{table*}[t]
\caption{Comparative results (\%) of GZSL on four datasets with ResNet-101 features under the PS protocol.} 
\label{tab:PS_GZSL}
\begin{center}
\begin{small}
\begin{sc}
\begin{tabular}{c|c|ccc|ccc|ccc|ccc}
    \hline \hline
    \multirow{2}*{\textbf{Type}}&\multirow{2}*{\textbf{Method}}&\multicolumn{3}{c|}{\textbf{aPY}} &\multicolumn{3}{c|}{\textbf{AwA}} &\multicolumn{3}{c|}{\textbf{SUN}} &\multicolumn{3}{c}{\textbf{CUB}} \\
    & &$Acc_{\mathcal{U}}$ & $Acc_{\mathcal{S}}$ & H &$Acc_{\mathcal{U}}$ & $Acc_{\mathcal{S}}$ & H &$Acc_{\mathcal{U}}$ & $Acc_{\mathcal{S}}$ & H &$Acc_{\mathcal{U}}$ & $Acc_{\mathcal{S}}$ & H \\
     \hline
    \multirow{9}*{Ind.} & ESZSL~\cite{romera2015embarrassingly} &2.4 &70.1& 4.6&6.6& 75.6& 12.1&11.0& 27.9& 15.8&12.6& 63.8& 21.0\\
    & SynC~\cite{changpinyo2016synthesized}& 7.4 &66.3& 13.3& 8.9 &$\bm{87.3}$ &16.2&7.9 &43.3 &13.4&11.5& \textcolor{blue}{70.9} &19.8\\
     & SAE~\cite{kodirov2017semantic} & 0.4& $\bm{80.9}$ &0.9&1.8 &77.1& 3.5& 8.8& 18.0 &11.8&7.8 &54.0 &13.6\\ 
    &CAPD~\cite{rahman2018unified} & \textcolor{blue}{26.8}&59.5&\textcolor{blue}{37.0}&45.2&68.6&54.5&35.8&27.8&31.3
    &44.9&41.7&43.3\\
    &f-CLSWGAN~\cite{xian2018feature} &-&-&-&\textcolor{blue}{57.9} &61.4 &59.6&\textcolor{blue}{42.6} &36.6 &\textcolor{blue}{39.4}&43.7 &57.7 &49.7\\
    &CDL~\cite{jiang2018learning}&19.8&48.6&28.1&28.1&73.5&40.6&23.5&$\bm{55.2}$&32.9
    &21.5&34.7&26.5\\
    & DCN~\cite{liu2018generalized} & 14.2 &\textcolor{blue}{75.0} &23.9 & 25.5 & \textcolor{blue}{84.2} & 39.1& 25.5 &37.0& 30.2 & 28.4 & 60.7 & 38.7\\
    &SE-ZSL~\cite{verma2018generalized}&-&-&-&56.3& 67.8& \textcolor{blue}{61.5}&40.9&30.5& 34.9
    &41.5& 53.3& 46.7\\
    &PreseR~\cite{annadani2018preserving} & 13.5&51.4&21.4&-&-&-&20.8&37.2&26.7 &24.6&54.3&33.9\\
    \hline
    \multirow{5}*{Trans.}
    &ALE~\cite{akata2016label}&-&-&9.6&-&-&26.1&-&-&21.5&-&-&31.5\\
    &GFZSL~\cite{verma2017simple}&-&-&0.0&-&-&48.5&-&-&0.0&-&-&33.1\\
    &DSRL~\cite{ye2017zero}&-&-&11.6&-&-&22.5&-&-&20.6 &-&-& 24.3\\
    &QFSL~\cite{song2018transductive}&-&-&-&-&-&-&$\bm{51.3}$&31.2
    &38.8&$\bm{71.5}$&$\bm{74.9}$&$\bm{73.2}$\\
   \hline
    &HPL  &$\bm{31.1}$ & 64.4 & $\bm{41.9}$ & $\bm{60.3}$ &75.7 & $\bm{67.1}$ & 39.1 & \textcolor{blue}{50.9} &$\bm{44.2}$ & \textcolor{blue}{47.1} & 54.2 & \textcolor{blue}{50.4} \\
    \hline\hline
    \end{tabular}
\end{sc}
\end{small}
\end{center}
\end{table*}

\textbf{Generalized ZSL.}
In real applications, whether a sample is from a seen or unseen class is unknown in advance. Hence, GZSL is a more practical and challenging task compared with standard ZSL. Here, we further evaluate the proposed model under the GZSL setting with PS protocol~\footnote{Note that GZSL has less been performed under the SS protocol due to its unreasonable data split.}.
The other experimental settings are kept the same as those used in~\cite{xian2018zero}, where the 2048-dim ResNet-101 features are adopted as the input, and compared approaches are consistent with those in Table~\ref{tab:PS_SZSL}.
The comparative results are shown in Table~\ref{tab:PS_GZSL}, much lower than those in standard ZSL. This is not surprising since the seen classes are included in the search space which act as distractors for the samples that come from unseen classes. 
Additionally, it can be observed that generally our method improves the overall performance (i.e., harmonic mean $H$) over the strongest competitor by an obvious margin (4.8\% $\sim$ 5.6\%). Such promising performance boost mainly comes from the improvement of mean class accuracy on the unseen classes, meanwhile without much performance degradation on the seen classes. These compelling results aslo verify that our method can significantly alleviate the strong bias towards seen classes by using the test samples from unseen classes. This is mainly due to the fact that, unlike most of existing transductive approaches (e.g., DIPL~\cite{zhao2018domain}) that only rely on projection learning, our HPL further introduces structural consistency constraint in the unseen class domain for transductive ZSL. 
Conversely, benefiting from the convincing predicted labels, the prototypes in both seen and unseen class domains are learned more discriminatively.

\begin{figure}[ht]
\begin{center}
\centerline{\includegraphics[width=3.5in]{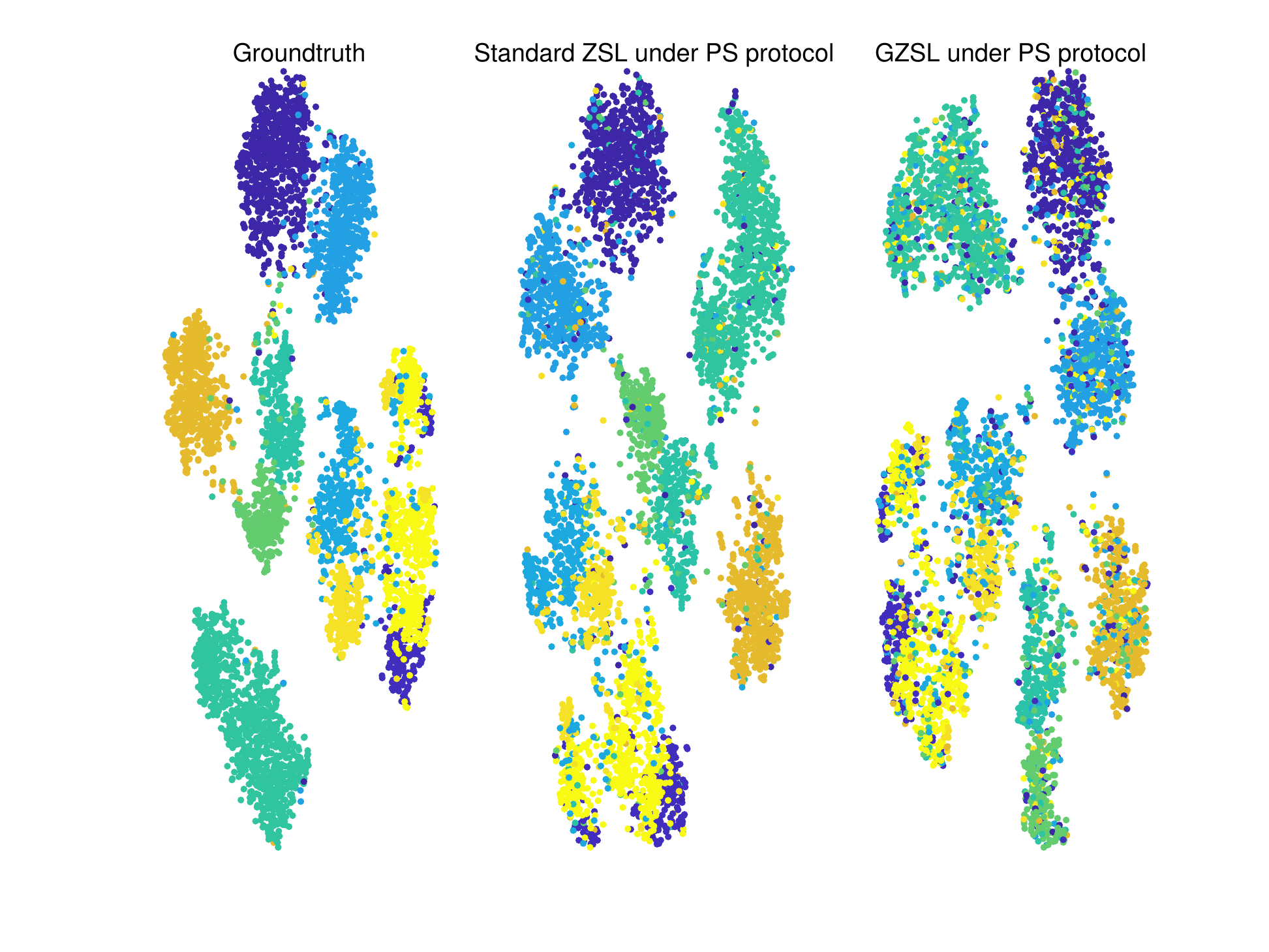}}
\caption{The t-SNE of all unseen-class data on AwA dataset with the true and predicted labels for standard ZSL and GZSL.}
\label{fig:tsne}
\end{center}
\vskip -0.2in
\end{figure}

For a straightforward illustration of our HPL in zero-shot recognition, we further show the t-SNE of all unseen-class data on AwA dataset with the true and predicted labels for standard ZSL and GZSL . As observed from Fig.~\ref{fig:tsne}, our HPL still can capture the global distribution of original data under various settings, although its performance under the GZSL setting is not as attractive as that under standard ZSL setting.


\subsection{Parameter Analysis}
\begin{figure*}[tbp]
\centering
\subfigure[$\rho$]{
\begin{minipage}[t]{0.25\linewidth}
\centering
\includegraphics[width=1.75in]{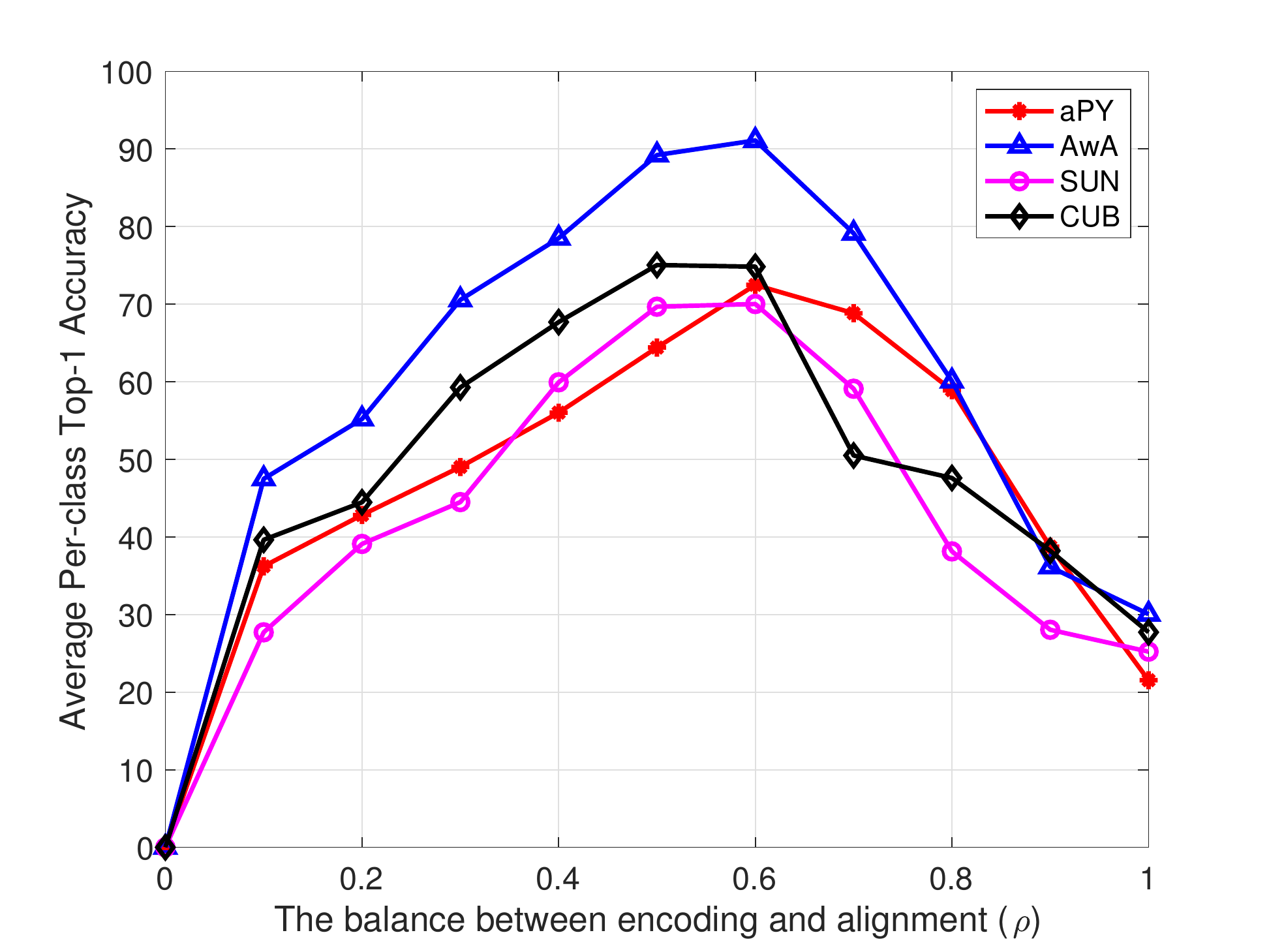}
\end{minipage}%
}%
\subfigure[$\omega$]{
\begin{minipage}[t]{0.25\linewidth}
\centering
\includegraphics[width=1.75in]{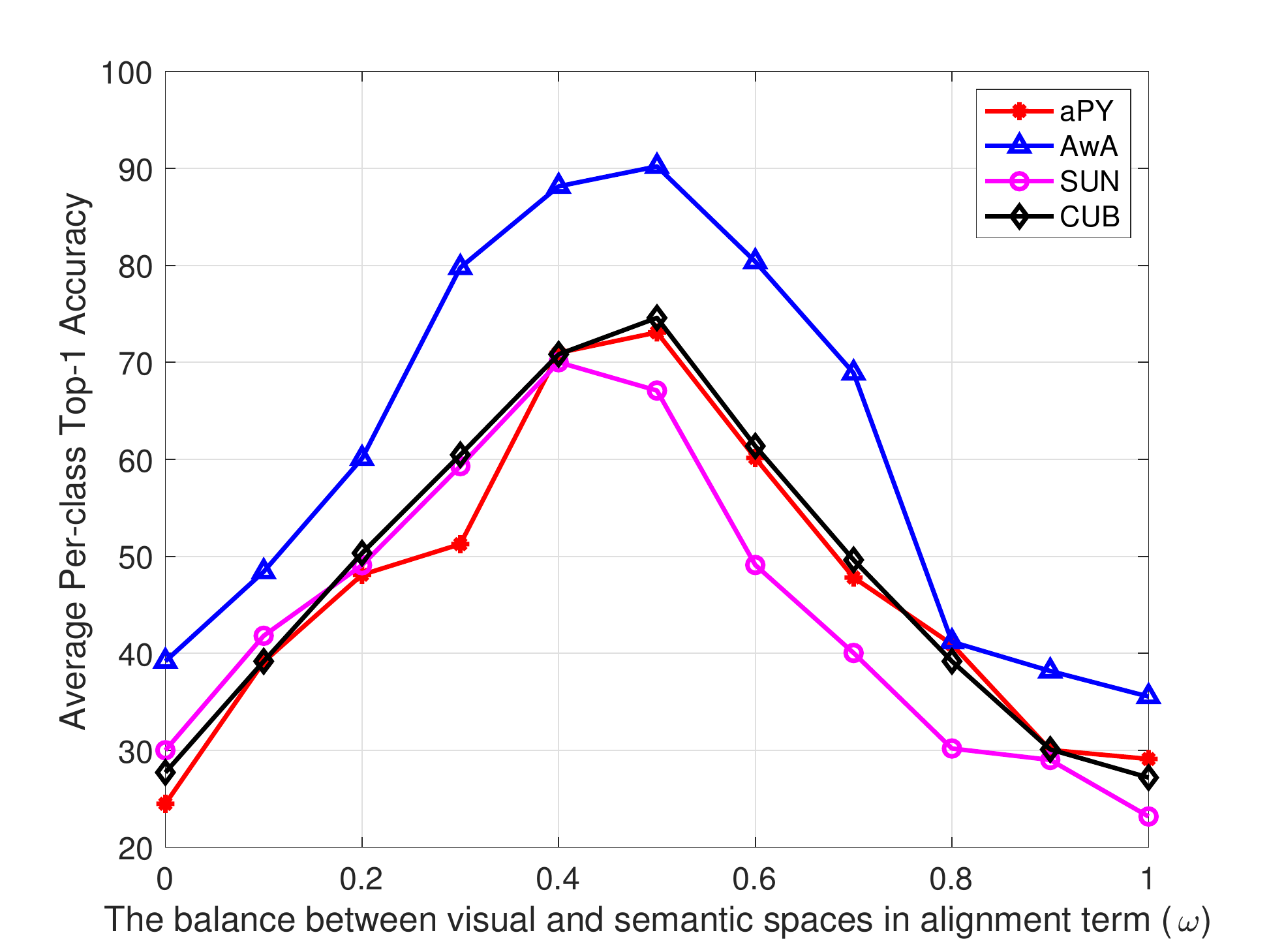}
\end{minipage}%
}%
\subfigure[$\alpha$]{
\begin{minipage}[t]{0.25\linewidth}
\centering
\includegraphics[width=1.75in]{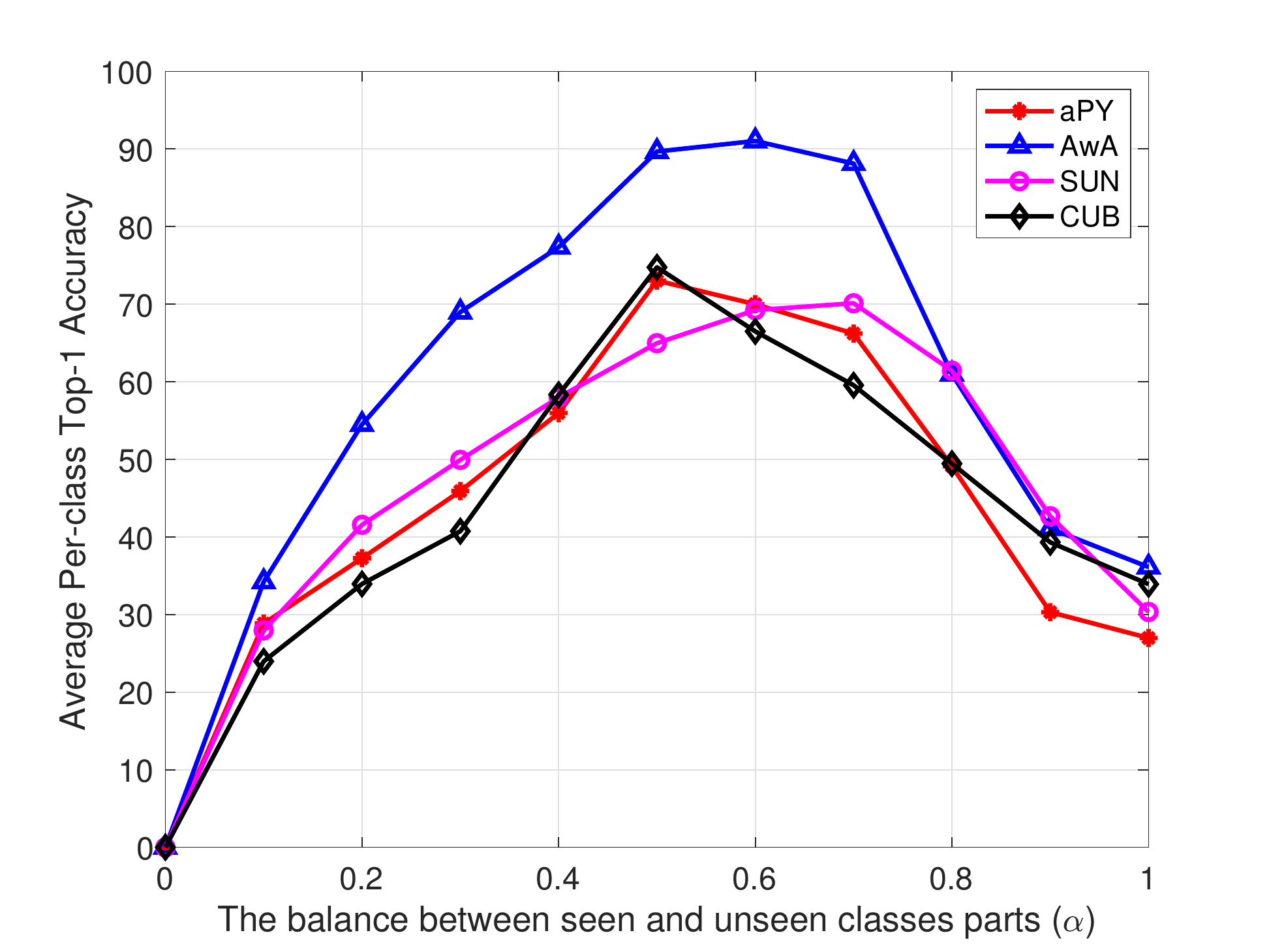}
\end{minipage}
}%
\subfigure[$\theta$]{
\begin{minipage}[t]{0.25\linewidth}
\centering
\includegraphics[width=1.75in]{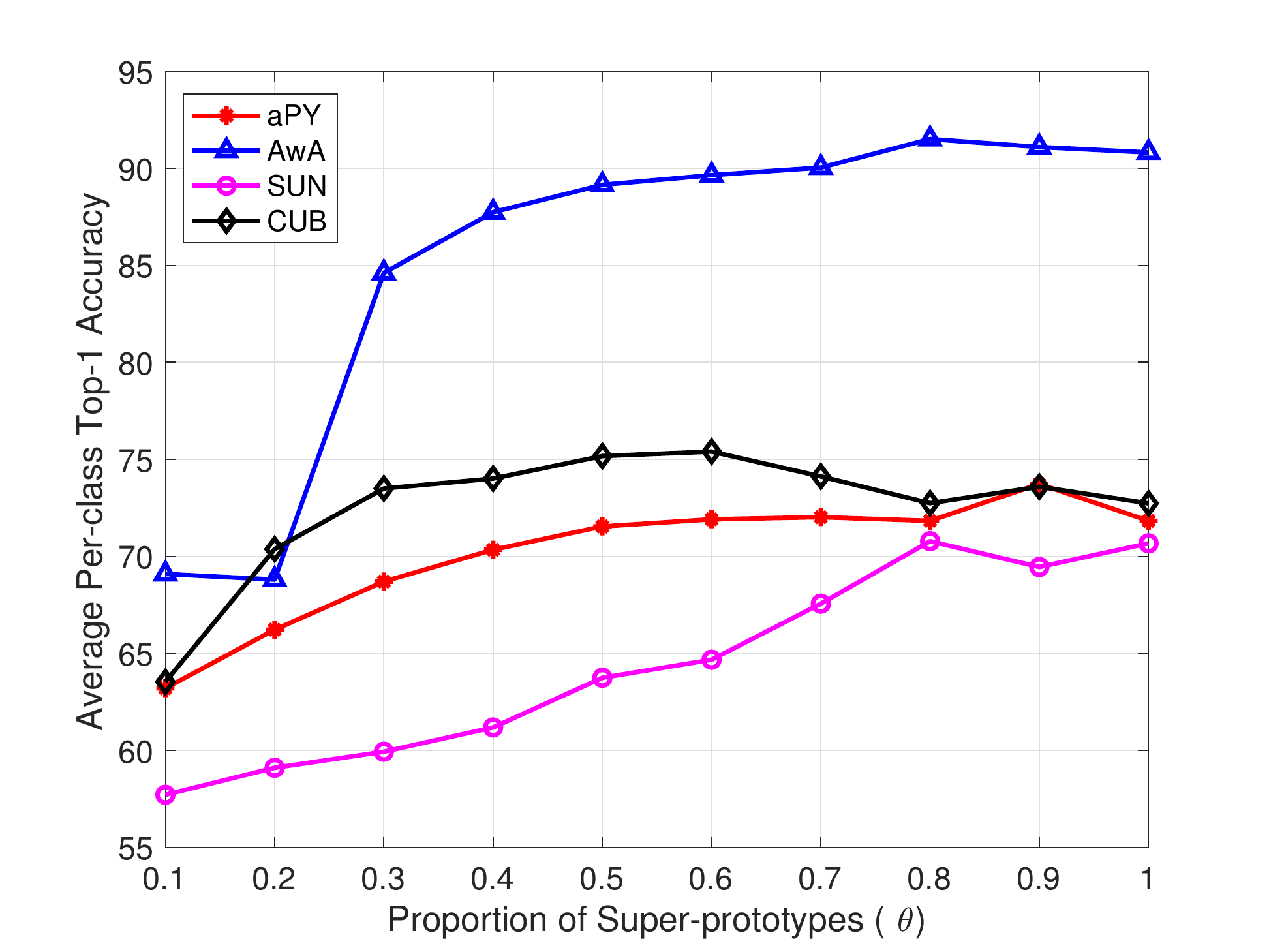}
\end{minipage}
}%
\centering
\vspace{-0.1in}
\caption{Standard ZSL performance as a function of the specific hyper-parameter $\left (\rho,\omega, \alpha, \theta \right )$ on the four datasets with ResNet-101 features under PS protocol.}
\label{fig:para}
\vspace{-0.1in}
\end{figure*}

It is worth noting that parameters $\left (\rho,\omega,\alpha \right )$ in the proposed model can be easily fine-tuned in the range $\left [ 0,1 \right)$ using the train and validation splits provided by~\cite{xian2018zero}, while it is not trivial for the parameter $q \in \{1,\cdots,m+n\}$ (i.e., the number of super-prototypes). This is because the optimal number of super-prototypes varies with the total number of classes. Additionally, when the seen and unseen classes are changed, the structure of super-prototypes will also be affected. To alleviate this issue, instead of $q$, we fine-tune the parameter $\theta =\frac{q}{m+n}\in[0,1]$, i.e., the proportion of super-prototypes to the total class number, using the validation set. 
\textcolor{black}{Fig.~\ref{fig:para} shows the effects of four parameters $\left (\rho,\omega,\alpha,\theta \right )$ in our HPL model respectively, where the standard ZSL task is performed on the four datasets under the PS protocol as in Table~\ref{tab:PS_SZSL}. It can be observed that generally, $\rho,\omega,\alpha \in [0.4, 0.7]$ and $\theta=\frac{m}{m+n}$ are enough to achieve promising recognition accuracy. Specifically, i) $\rho$ peaks around 0.6, which means the visual encoding cost should overweigh the alignment term slightly. This is reasonable since our zero-shot recognition depends on the encoding term directly while the alignment term actually serves as a regularizer. ii) $\omega$ peaks around 0.5 since the visual and semantic spaces should be equally important in alignment term. iii) $\alpha$ peaks around 0.6, where the accuracy would first increase and then decrease along with increasing the value of $\alpha$, corresponding to overfitting and under-fitting of our model on seen-class training sets. iv) $\theta=\frac{m}{m+n}$ means the optimal value of $q$ is approximate to $m$ (the number of seen classes), since $m$ is often larger than $n$ (the number of unseen classes).}

\begin{table*}[!t]
\caption{Comparative results of standard ZSL on four datasets with ResNet-101 features under the PS protocol.} 
\label{tab:compare_ablation}
\begin{center}
\begin{sc}
\begin{tabular}{c|c|c|cccc}
    \hline \hline
    \textbf{Type} & \textbf{Ablation Index} &\textbf{Algorithm} & \textbf{aPY} & \textbf{AwA} & \textbf{SUN} & \textbf{CUB}  \\
    \hline
    \multirow{9}*{Accuracy(\%)}& \multirow{2}*{i}
    & HPL1 &72.2 & \textbf{91.9}& 68.3 & \textbf{75.4} \\
    & & HPL  &\textbf{73.8} & 91.2 & \textbf{70.4} & 75.2 \\
    \cline{2-7}
    & \multirow{2}*{ii}
    & Seen classes part first  &72.9 & 91.0 & 69.4 & \textbf{75.7} \\
    & & Unseen classes part first  &\textbf{73.8} & \textbf{91.2} & \textbf{70.4} & 75.2 \\
    \cline{2-7}
    & \multirow{3}*{iii}
    & Kmeans~\cite{duda2012pattern}  &73.2 & 90.2 & 68.2 & \textbf{75.5} \\
    & & AP~\cite{frey2007clustering}  &71.1 & \textbf{91.4} & 68.8 & 74.1 \\
    & & Each Class Mean  &\textbf{73.8} & 91.2 & \textbf{70.4} & 75.2 \\
    \cline{2-7}
    & \multirow{2}*{iv}
    & without alternative  &67.8 & 85.9 & 65.0 & 67.5 \\
    & & with alternative     &\textbf{73.8} & \textbf{91.2} & \textbf{70.4} & \textbf{75.2} \\
    \hline
    \multirow{9}*{Time(sec.)}& \multirow{2}*{i}
    & HPL1& 1091.0 & 697.3 & 1101.6 & 631.6 \\
    & & HPL & \textbf{702.7} & \textbf{490.9} & \textbf{811.2}  & \textbf{333.4} \\
    \cline{2-7}
    & \multirow{2}*{ii}
    & Seen classes part first & 1034.2 & 592.1 & 1230.1 & 572.1 \\
    & & Unseen classes part first  & \textbf{702.7} & \textbf{490.9} & \textbf{811.2}  & \textbf{333.4} \\
    \cline{2-7}
    & \multirow{3}*{iii}
    & Kmeans~\cite{duda2012pattern} & 1101.1 & 892.1 & \textbf{782.1} & \textbf{308.3} \\
    & & AP~\cite{frey2007clustering} & 981.9 & 689.6 & 992.3 & 482.5 \\
    & & Each Class Mean  & \textbf{702.7} & \textbf{490.9} & 811.2  & 333.4 \\
    \cline{2-7}
    & \multirow{2}*{iv}
    & without alternative & \textbf{167.1} & \textbf{89.4} & \textbf{201.5} & \textbf{68.7} \\
    & & with alternative & 702.7 & 490.9 & 811.2  & 333.4 \\
    \hline\hline
    \end{tabular}
\end{sc}
\end{center}
\end{table*}

\subsection{Ablation Study}
\textcolor{black}{
To verify the advantage of the proposed HPL model and optimization rules in Algorithm~\ref{alg:Framwork}, we additionally conduct four ablation studies on all datasets with Resnet-101 features under the PS protocol of standard ZSL as follows.
i) The bidirectional projection is also introduced in Eq.~(\ref{Super_Prototype}) and Eq.~(\ref{Unseen_alig}) as in Eq.~(\ref{Seen_Prototype}) and Eq.~(\ref{Unseen_enc}), dubbed \text{HPL1}. 
This will influence the update of all variables except $\bm C_{u}$, but the optimization strategy about each variable is the same as that for our HPL model. Table~\ref{tab:compare_ablation} reports the accuracy and physical running time of two models. It can be observed that HPL1 obtains similar accuracy to our HPL but spends more time. 
Thus, it is generally enough to consider the bidirectional projection only in Eq.~(\ref{Seen_Prototype}), since Eq.~(\ref{Super_Prototype}) actually serves as a regularizer of Eq.~(\ref{Seen_Prototype}).
}

\textcolor{black}{
For Algorithm~\ref{alg:Framwork}, ii) we exchange the steps in lines 5 and 6 to update the seen classes part first. The comparison results are presented in Table~\ref{tab:compare_ablation}. It can be concluded that optimizing unseen classes part first can achieve similar and even more promising accuracy to the case of seen classes part first, and meanwhile, take much less time.
iii) Meanwhile, to analyze the initialization sensitivity of $\bm D_{v}^{(0)}$ and $\bm D_{c}^{(0)}$, we now additionally initialize $\bm P_{s}$ and $\bm P_{u}$ using Kmeans~\cite{duda2012pattern} and AP~\cite{frey2007clustering}. As compared in Table~\ref{tab:compare_ablation}, we find that different initial values have no obvious effects on ZSL accuracy, while slight effects on physical running time of our algorithm. This is reasonable since different initialization conditions generally influence the convergence rate of alternative optimization algorithm.}
iv) Finally, we evaluate the effect of proposed alternative optimization rules in Algorithm~\ref{alg:Framwork}. Specifically, we first train super-prototypes $\bm D_{v}$ and $\bm D_{c}$ from seen classes via Eq.~(\ref{ObjectApproximate1}) with $\gamma =0$, and then employ super-prototypes to the unseen classes to solve Eq.~(\ref{Unseen}). By comparison from Table~\ref{tab:compare_ablation}, we can find that the standard ZSL accuracy by alternative optimization algorithm is considerably superior to that without alternative update, though takes a little more time. Additionally, the importance of each component in our HPL model can be observed from parameter analysis results in Fig.~\ref{fig:para}. Obviously, $\rho$ (i.e., $\beta$) peaks around 0.6 instead of 0, which validates the encoding function makes a little more effect on final performances than the alignment function. The bidirectional projection in encoding function is also necessary since $\omega$ (i.e., $\lambda$) peaks around 0.5. In particular, $\alpha$ (i.e., $\gamma$) peaks around 0.6 instead of 0, which shows employing test data in our model is indeed beneficial to ZSL.

\section{Conclusions and Future Work}
In this paper, we propose a hierarchical prototype learning model (HPL) that is able to perform efficient zero-shot recognition in the original visual space, and meanwhile, avoid a series of problems caused by the provided semantic prototypes. In particular, the discriminability of visual prototypes is further strengthened by coupling them with semantic prototypes in an aligned space, thus achieving more promising recognition performance. Furthermore, interpretable super-prototypes shared between the seen and the unseen class domains are exploited to alleviate the domain shift issue. We have carried out extensive experiments about ZSL on five benchmarks, and the results demonstrate the obvious superiority of the proposed HPL to the state-of-the-art approaches. 
It is also worth noting that the number of visual/semantic prototypes is not controllable in our HPL. In essence, learning one prototype for a class is generally insufficient to recognize one class and differentiate two classes. Thus, our ongoing research work includes learning prototypes adaptively with the data distribution.





\ifCLASSOPTIONcaptionsoff
  \newpage
\fi




\bibliographystyle{IEEEtran}
\bibliography{mybibfile_new}

\end{document}